\documentclass[twocolumn]{bytedance_seed}

\usepackage[toc,page,header]{appendix}

\usepackage{minitoc}
\usepackage{graphicx}
\usepackage{amsmath}
\usepackage{amssymb}
\usepackage{mathtools}
\usepackage{amsthm}
\usepackage{hyperref}
\usepackage{multirow}
\usepackage{xcolor}
\usepackage{colortbl}
\usepackage{booktabs}

\usepackage{algorithm}
\usepackage{algorithmicx}
\usepackage{algpseudocode}
\usepackage{verbatim}
\usepackage{caption}
\usepackage{subcaption}
\usepackage{fancyvrb}
\usepackage{fvextra}
\usepackage{wrapfig}
\usepackage{xurl}

\definecolor{GREEN}{HTML}{62b197}
\definecolor{RED}{HTML}{e18e6d}

\DeclareUnicodeCharacter{FF5C}{\textbar}

\title{Feedback-Driven Tool-Use Improvements in Large Language Models via Automated Build Environments}

\author[1,2,*]{Junjie Ye}
\author[1]{Changhao Jiang}
\author[2]{Zhengyin Du}
\author[2]{Yufei Xu}
\author[2]{Xuesong Yao}
\author[1]{\\Zhiheng Xi}
\author[1]{Xiaoran Fan}
\author[1]{Qi Zhang}
\author[1]{Tao Gui}
\author[1]{Xuanjing Huang}
\author[2]{Jiecao Chen}

\affiliation[1]{Fudan University}
\affiliation[2]{ByteDance Seed}

\contribution[*]{Work done at ByteDance Seed}

\abstract{
Effective tool use is essential for large language models (LLMs) to interact with their environment. However, progress is limited by the lack of efficient reinforcement learning (RL) frameworks specifically designed for tool use, due to challenges in constructing stable training environments and designing verifiable reward mechanisms.
To address this, we propose an automated environment construction pipeline, incorporating scenario decomposition, document generation, function integration, complexity scaling, and localized deployment. This enables the creation of high-quality training environments that provide detailed and measurable feedback without relying on external tools.
Additionally, we introduce a verifiable reward mechanism that evaluates both the precision of tool use and the completeness of task execution. When combined with trajectory data collected from the constructed environments, this mechanism integrates seamlessly with standard RL algorithms to facilitate feedback-driven model training.
Experiments on LLMs of varying scales demonstrate that our approach significantly enhances the models’ tool-use performance without degrading their general capabilities. Our analysis suggests that these gains result from improved context understanding and reasoning, driven by updates to the lower-layer MLP parameters in models.
}

\correspondence{Junjie Ye at \email{jjye23@m.fudan.edu.cn}}

\checkdata[Code \& Data]{\url{https://github.com/bytedance/FTRL}}

\begin{document}
\maketitle


\section{Introduction}

\begin{figure}[!t]
    \centering
    \includegraphics[width=\linewidth]{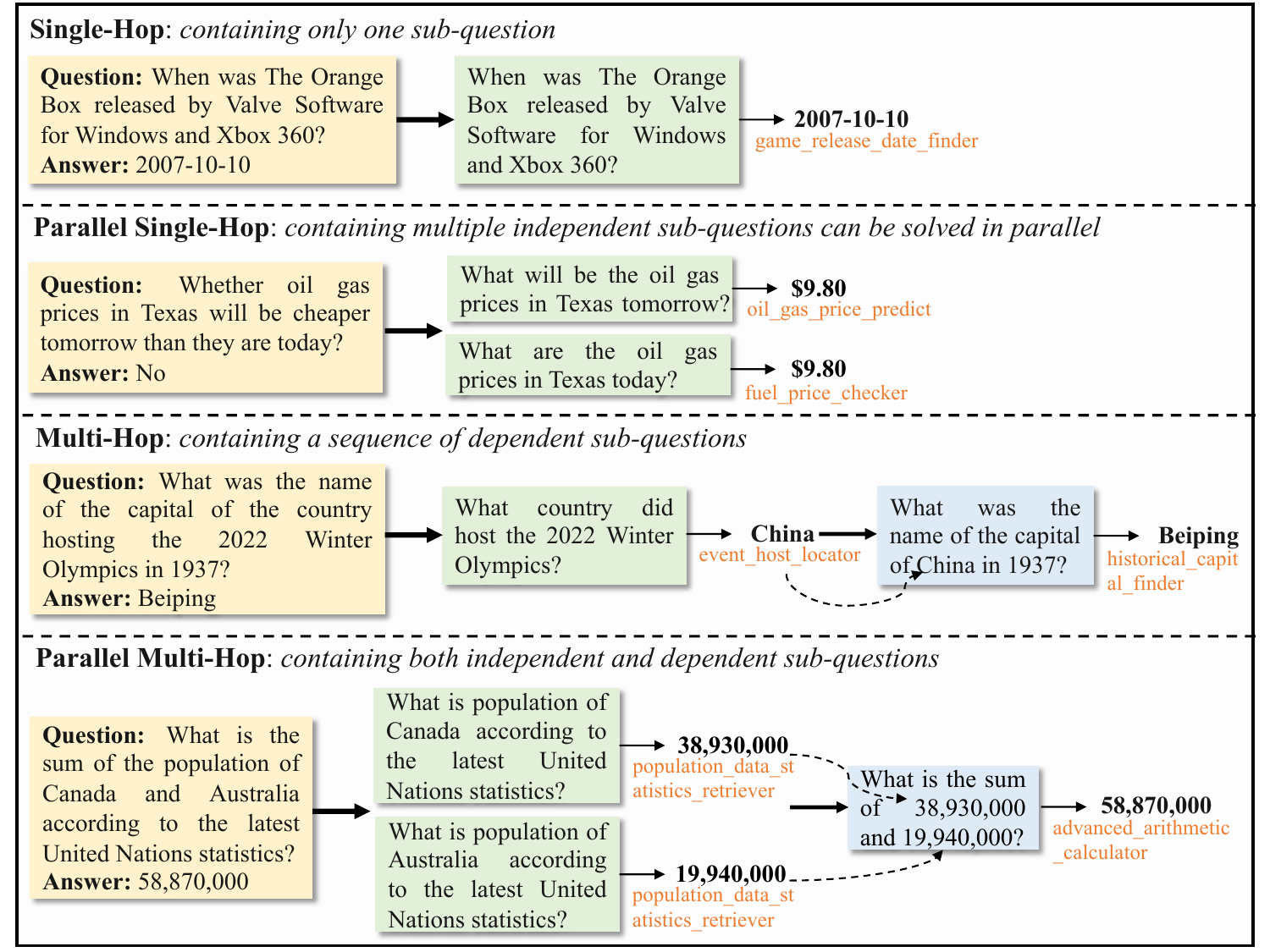}
    \caption{Illustrative examples of four scenarios, categorized by varying sub-question pattern combinations.}
    \label{fig:example}
\end{figure}

Tool use in large language models (LLMs)~\cite{Claude, GPT-4, Gemini1.5} refers to their ability to interact with the external world by invoking tools to retrieve information~\cite{Toolalpaca} or respond to environmental stimuli~\cite{ToolSword}. Consequently, the effectiveness of an LLM's tool-use capabilities significantly impacts its performance on complex real-world tasks~\cite{tool-learning, OpenNovelty} and represents a key step toward achieving general-purpose AI~\cite{agent-survey}.

Recent research has focused on improving LLM tool-use capabilities through post-training methods~\cite{tool-survey}. A common approach involves fine-tuning open-source models on interaction trajectories generated by proprietary LLMs with diverse APIs~\cite{Toolformer, Toolalpaca}. To increase the diversity and quality of these trajectories, various techniques have been proposed, including Monte Carlo sampling~\cite{Agent-R}, graph-based inference~\cite{ControlLLM}, and other optimization strategies~\cite{ToolChain}. In parallel, reinforcement learning (RL) methods have been explored to improve adaptability and decision-making through active environment interaction~\cite{StepTool, ToolRL, TL-Training}.

However, current RL-based frameworks for training tool-use face significant limitations that hinder the development of robust models. One major challenge is the difficulty in constructing stable training environments. These frameworks typically rely on a large set of online tools, which are prone to service disruptions due to factors like API rate limits~\cite{RestGPT, ToolQA}. Additionally, standardizing the organization and deployment of these tools imposes high infrastructure costs~\cite{Toolllm}. Another key limitation is the lack of verifiable reward signals. The complexity of tool interactions and the diversity of valid trajectories often necessitate the use of high-level LLMs for evaluation, which introduces model bias and undermines both training efficiency and algorithmic stability~\cite{StepTool, ToolRL}.

To address these limitations, we propose an automated pipeline for generating a large number of high-quality tool-use training environments. This pipeline includes scenario decomposition, document generation, function integration, complexity scaling, and localized deployment. By executing all tools locally as code, it enables the dynamic creation of diverse, stable environments while ensuring controlled feedback, free from reliance on external online toolsets. Furthermore, each environment is built around a well-defined set of target sub-questions, allowing for precise evaluation of model behavior.

Building on this environment, we introduce a verifiable reward mechanism that evaluates the precision of tool use and the completeness of task execution solely by analyzing feedback from the environment. When combined with interaction trajectory data collected, this mechanism can be seamlessly integrated into any preference-based optimization method. Together, these components form a feedback-driven framework for training LLMs in robust and effective tool use.

We validate our approach through extensive experiments on LLMs of varying sizes. Results show that our method consistently improves tool-use performance across four distinct benchmarks, while preserving the models’ general capabilities, regardless of the model architecture, inference patterns, or training algorithms employed. A parameter-level analysis suggests that these gains are largely driven by updates in lower-layer MLP parameters, which enhance the model’s contextual understanding and decision-making.

In summary, our main contributions are:
\begin{itemize}
    \item We propose an automated environment construction pipeline that ensures stable and verifiable tool-use training;
    \item We introduce a verifiable reward mechanism that jointly evaluates precision and completeness, compatible with preference-based optimization strategies;
    \item We conduct extensive experiments demonstrating substantial improvements in tool use, supported by parameter-level analysis.
\end{itemize}

\section{Related Work}

\paragraph{Methods for Designing Tool-Use Environments}
The construction of diverse tool-use environments is both necessary and critical for advancing research. Early studies primarily focused on a limited set of predefined tools, for which large volumes of data were either manually curated or synthetically generated within narrowly defined scenarios~\cite{Toolformer, GPT4tool, metatool}. To support more diverse tool-use tasks, later efforts collected various existing APIs from the web and organized them into large toolsets. Synthetic user inputs were then generated to evaluate the performance of LLMs~\cite{RestGPT, Toolllm, ToolChain, ToolEyes}. However, these approaches remain constrained by the quality and coverage of the collected toolsets and often suffer from issues such as unstable execution environments and unreliable feedback. To address these limitations,~\citet{ToolHop} introduced a controllable method for constructing multi-hop tool-use environments. While this represented progress, their approach was restricted to the multi-hop scenario and focused solely on building test data. In contrast, we propose a robust automated framework that supports the generation of diverse tool-use scenarios and enables both effective training and evaluation, thereby providing a more comprehensive foundation for research in this domain.

\begin{figure*}[!t]
    \centering
    \includegraphics[width=\linewidth]{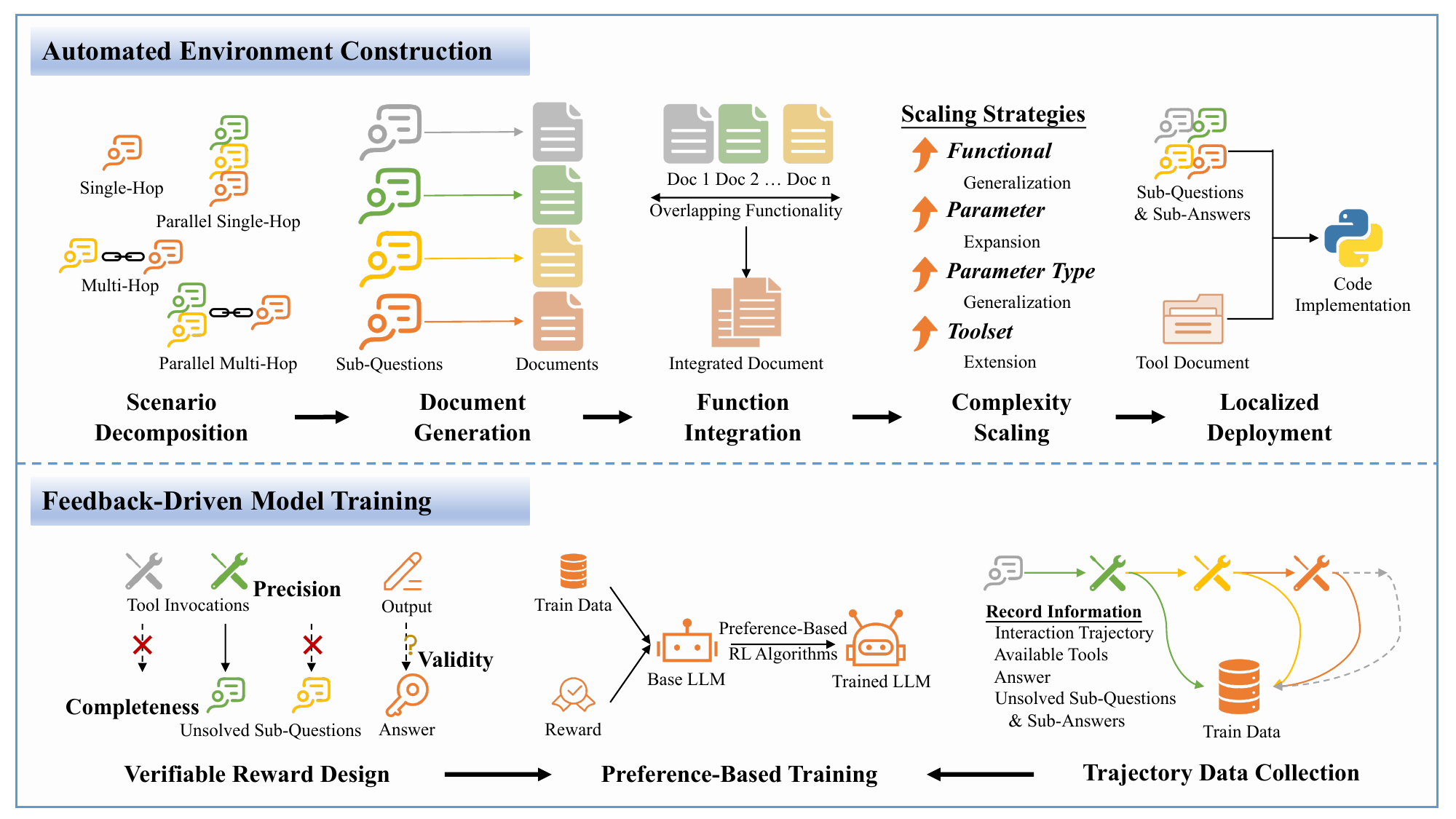}
    \caption{Overview of FTRL. The automated environment construction follows a five-stage pipeline to generate diverse tool-use training environments. Feedback-driven model training then collects data within these environments, incorporates verifiable reward mechanisms, and optimizes performance using preference-based RL algorithms.}
    \label{fig:approach}
\end{figure*}

\paragraph{Techniques for Enhancing Tool-Use Abilities}
To improve the tool-use capabilities of LLMs, early research typically relied on a limited number of predefined tool environments. These efforts either used in-context learning to prompt the model to invoke tools~\cite{ART} or manually constructed large datasets of tool-use instances for supervised training~\cite{document}. To further enhance tool comprehension, some approaches directly encoded tools as part of the model’s vocabulary during training~\cite{Toolken, ToolGen}. To encourage generalization to out-of-domain tools, many existing studies format tool document using a unified JSON Schema and train models to perform tool calls using structured formats such as ReAct~\cite{React} or CodeAct~\cite{CodeAct}. With advances in RL, more recent work has explored enabling models to improve tool-use capabilities through interaction with their environments~\cite{StepTool, ToolRL, TL-Training}. However, these approaches have yet to deliver a universal training framework, primarily due to the difficulty of efficiently constructing diverse and stable training environments, and the lack of reliable reward signals for guiding learning. In contrast, we introduce a feedback-driven training scheme grounded in the automated construction of diverse tool-use environments. Combined with a verifiable reward mechanism, our approach effectively enhances the tool-use capabilities of LLMs using only feedback from a stable environment.

\section{Approaches}

As illustrated in Figure~\ref{fig:approach}, our approach comprises two core components: 1) a five-stage automated pipeline for constructing tool-use training environments for LLMs; and 2) a feedback-driven model training framework that leverages these environments to enhance tool-use capabilities.

\subsection{Automated Environment Construction}
\label{sec:env}

A complete training environment for tool use consists of a user input $\mathbb{U}$ and a toolset $\mathbb{T}$. $\mathbb{U}$ contains a question $q$ and its corresponding answer $a$, where $q$ can be decomposed into a sequence of sub-questions $q_1, q_2, \ldots, q_n$. Each sub-question $q_i$ is solvable via a tool from $\mathbb{T}$, producing intermediate answers $a_i$.
Each tool $t_i \in \mathbb{T}$ is specified by a document $d_i$ and an implementation component $c_i$. The document $d_i$ defines the tool's name $n_i$, function description $f_i$, and a set of parameters $\mathbb{P}_i$. Each parameter $p_{ij} \in \mathbb{P}_i$ is defined by its name $pn_{ij}$, type $pt_{ij}$, description $pf_{ij}$, and a binary flag $pr_{ij}$ indicating whether it is required.
To efficiently construct robust training environments for tool use, we propose a five-stage automated environment construction pipeline.

\paragraph{Scenario Decomposition}
To ensure the diversity of training environments, we account for different logical relationships among sub-questions, defining four scenarios of tool use, as illustrated in Figure~\ref{fig:example}:
1) Single-hop, containing only one sub-question $q_1$;
2) Parallel single-hop, where $q$ is decomposed into multiple independent sub-questions $q_1, q_2, \ldots, q_n$ that can be solved in parallel;
3) Multi-hop, where $q$ is broken down into a sequence of dependent sub-questions, such that each $q_{i+1}$ relies on the answer $a_i$ of the preceding sub-question; 
and 4) Parallel multi-hop, a hybrid structure consisting of both independent and dependent sub-questions.
We manually construct user inputs for each scenario to ensure broad coverage and task variability.

\paragraph{Document Generation}
After constructing diverse user inputs, we ensure their solvability by generating a corresponding tool document $d_i$ for every sub-question $q_i$. For every sub-question $q_i$, we generate a tool document $d_i$ with a function $f_i$ explicitly designed to solve the task described by $q_i$. The parameter set $\mathbb{P}_i$ is also abstracted directly from $q_i$, establishing a precise one-to-one mapping between sub-questions and tool interfaces.

\paragraph{Function Integration}
While generating a separate tool for each sub-question ensures full coverage, it often results in redundancy within the toolset. To reduce duplication, we analyze tool documents $d_1, d_2, \ldots, d_n$ and merge those with overlapping functionality. The resulting set $d_1, d_2, \ldots, d_m$ ($m \leq n$) offers improved modularity and efficiency, while preserving logical consistency with the original task.

\begin{table}[!t]
    \centering
    \resizebox{\linewidth}{!}
    {
    \begin{tabular}{l c cccc}
    \toprule
    \multirow{2}*{\textbf{Datasets}}     & {\textbf{Train}} & \multicolumn{4}{c}{\textbf{Test}}  \\ \cmidrule(lr){2-2} \cmidrule(lr){3-6}
         & \textbf{Ours} & \textbf{Ours} & \textbf{ToolHop} & \textbf{$\tau$-bench} & \textbf{RoTBench} \\ \midrule
    \# Scenarios & 4 	& 4 	& 3	& 2 	& 5 \\
    \# Number & 2215	& 200	& 2985	& 165	& 840 \\
    \# Avg Tools & 9.39  	& 9.26 	& 3.93 	& 12.36 & 8.50 \\
    \bottomrule
    \end{tabular}
    }
    \caption{Statistics of datasets. `\# Scenarios' refers to the number of tool-use scenarios; `\# Number' indicates the total number of data instances; `\# Avg Tools' denotes the average number of tools included in each instance.}
    \label{tab:datasets}
\end{table}

\begin{table*}[!t]
    \centering
    \resizebox{\linewidth}{!}{
    \begin{tabular}{ll lll l l lll l}
    \toprule
    \multirow{2}*{\textbf{Family}} & \multirow{2}*{\textbf{Version}} & \multicolumn{3}{l}{\textbf{Ours}} & \textbf{ToolHop} & \textbf{$\tau$-bench} & \multicolumn{3}{l}{\textbf{RoTBench}} & \multirow{2}*{\textbf{Avg}} \\ 
    \cmidrule(l){3-5} \cmidrule(l){6-6} \cmidrule(l){7-7} \cmidrule(l){8-10} 
     & & \textbf{Solve-P} & \textbf{Solve-R} & \textbf{Solve-F1} & \textbf{AC} & \textbf{Pass\texttt{\char`^}1} & \textbf{TS} & \textbf{PI} & \textbf{CF} & \\    
    \midrule
    \multicolumn{11}{c}{\textit{Closed-Source LLMs}} \\
    {Gemini} & 2.5-Flash & \textbf{52.78} & 23.13 & 23.24 & 39.37 & 40.04 & 53.62 & 34.95 & 22.76 & 36.24 \\
     Gemini& 2.5-Pro & \underline{51.77} & 31.96 & 29.98 & 45.32 & \underline{47.09} & 49.05 & 42.38 & 27.52 & 40.63 \\
    Claude & 4.0-Sonnet & 42.33 & 56.07 & 42.59 & \textbf{53.97} & \textbf{50.22} & 39.72 & 34.95 & 21.81 & 42.71 \\
    {GPT} & 3.5-Turbo & 25.77 & 29.29 & 25.66 & 25.73 & 15.13 & 65.71 & 29.81 & 17.33 & 29.30 \\
     GPT& 4o & 33.43 & 42.01 & 33.39 & 36.31 & 37.43 & \textbf{78.76} & \textbf{49.43} & \textbf{31.52} & 42.79 \\
    \midrule
    \multicolumn{11}{c}{\textit{Open-Source LLMs (32B - 72B)}} \\
    Qwen2.5-32B & Instruct & 30.12 & 37.80 & 32.42 & 19.70 & 21.91 & 75.43 & 36.86 & 20.00 & 34.28 \\
    Qwen2.5-72B & Instruct & 31.68 & 43.41 & 35.22 & 30.29 & 34.26 & 72.19 & 26.29 & 17.14 & 36.31 \\
    Qwen3-32B & Non-Reasoning & 33.20 & 46.24 & 36.34 & 35.34 & 27.39 & 73.24 & 41.71 & 24.86 & 39.79 \\
    Qwen3-32B & Reasoning & 28.94 & 40.41 & 31.30 & \underline{50.12} & 31.00 & 53.90 & 33.81 & 20.19 & 36.21 \\
    \midrule
    \multicolumn{11}{c}{\textit{Open-Source LLMs (7B - 14B)}} \\
    \rowcolor{gray!10} {Qwen2.5-7B} & Instruct & 27.44 & 27.69 & 25.97 & 11.99 & 5.91 & 70.38 & 27.33 & 15.43 & 26.52 \\
     & FTRL-Reinforce++ & 41.61 \textcolor{GREEN}{$\uparrow$} & 41.36 \textcolor{GREEN}{$\uparrow$} & 40.36 \textcolor{GREEN}{$\uparrow$} & 23.45 \textcolor{GREEN}{$\uparrow$} & 11.91 \textcolor{GREEN}{$\uparrow$} & 70.95 \textcolor{GREEN}{$\uparrow$} & 40.95 \textcolor{GREEN}{$\uparrow$} & 26.10 \textcolor{GREEN}{$\uparrow$} & 37.09 \textcolor{GREEN}{$\uparrow$} \\
     & FTRL-GRPO & 47.82 \textcolor{GREEN}{$\uparrow$} & 47.37 \textcolor{GREEN}{$\uparrow$} & 46.78 \textcolor{GREEN}{$\uparrow$} & 29.68 \textcolor{GREEN}{$\uparrow$} & 6.91 \textcolor{GREEN}{$\uparrow$} & 74.00 \textcolor{GREEN}{$\uparrow$} & 30.86 \textcolor{GREEN}{$\uparrow$} & 18.95 \textcolor{GREEN}{$\uparrow$} & 37.80 \textcolor{GREEN}{$\uparrow$} \\
    \rowcolor{gray!10} {Qwen2.5-14B} & Instruct & 33.31 & 34.64 & 32.76 & 25.80 & 16.74 & 71.81 & 38.29 & 21.33 & 34.33 \\
     & FTRL-Reinforce++ & 49.51 \textcolor{GREEN}{$\uparrow$} & 47.20 \textcolor{GREEN}{$\uparrow$} & 47.57 \textcolor{GREEN}{$\uparrow$} & 36.42 \textcolor{GREEN}{$\uparrow$} & 26.83 \textcolor{GREEN}{$\uparrow$} & 73.05 \textcolor{GREEN}{$\uparrow$} & \underline{45.24} \textcolor{GREEN}{$\uparrow$} & \underline{28.19} \textcolor{GREEN}{$\uparrow$} & 44.25 \textcolor{GREEN}{$\uparrow$} \\
     & FTRL-GRPO & 48.67 \textcolor{GREEN}{$\uparrow$} & 50.56 \textcolor{GREEN}{$\uparrow$} & 49.20 \textcolor{GREEN}{$\uparrow$} & 26.63 \textcolor{GREEN}{$\uparrow$} & 25.43 \textcolor{GREEN}{$\uparrow$} & 71.52 \textcolor{RED}{$\downarrow$} & 36.19 \textcolor{RED}{$\downarrow$} & 21.62 \textcolor{GREEN}{$\uparrow$} & 41.23 \textcolor{GREEN}{$\uparrow$} \\
    \rowcolor{gray!10} {Qwen3-8B} & Non-Reasoning & 21.18 & 30.71 & 23.48 & 28.54 & 10.13 & 75.52 & 36.29 & 22.19 & 31.01 \\
     & FTRL-Reinforce++ & 41.96 \textcolor{GREEN}{$\uparrow$} & 47.37 \textcolor{GREEN}{$\uparrow$} & 43.63 \textcolor{GREEN}{$\uparrow$} & 37.79 \textcolor{GREEN}{$\uparrow$} & 21.96 \textcolor{GREEN}{$\uparrow$} & \underline{78.10} \textcolor{GREEN}{$\uparrow$} & 42.57 \textcolor{GREEN}{$\uparrow$} & 25.90 \textcolor{GREEN}{$\uparrow$} & 42.41 \textcolor{GREEN}{$\uparrow$} \\
     & FTRL-GRPO & 46.44 \textcolor{GREEN}{$\uparrow$} & 55.84 \textcolor{GREEN}{$\uparrow$} & \underline{49.54} \textcolor{GREEN}{$\uparrow$} & 38.93 \textcolor{GREEN}{$\uparrow$} & 23.35 \textcolor{GREEN}{$\uparrow$} & 77.62 \textcolor{GREEN}{$\uparrow$} & 44.67 \textcolor{GREEN}{$\uparrow$} & 27.05 \textcolor{GREEN}{$\uparrow$} & \textbf{45.43} \textcolor{GREEN}{$\uparrow$} \\
    \rowcolor{gray!10} {Qwen3-14B} & Non-Reasoning & 28.29 & 37.58 & 29.97 & 24.19 & 13.74 & 75.90 & 36.38 & 20.67 & 33.34 \\
     & FTRL-Reinforce++ & 44.06 \textcolor{GREEN}{$\uparrow$} & \textbf{59.30} \textcolor{GREEN}{$\uparrow$} & 48.24 \textcolor{GREEN}{$\uparrow$} & 38.32 \textcolor{GREEN}{$\uparrow$} & 17.61 \textcolor{GREEN}{$\uparrow$} & 77.81 \textcolor{GREEN}{$\uparrow$} & 42.57 \textcolor{GREEN}{$\uparrow$} & 25.24 \textcolor{GREEN}{$\uparrow$} & 44.14 \textcolor{GREEN}{$\uparrow$} \\
     & FTRL-GRPO & 49.41 \textcolor{GREEN}{$\uparrow$} & \underline{56.71} \textcolor{GREEN}{$\uparrow$} & \textbf{51.82} \textcolor{GREEN}{$\uparrow$} & 38.26 \textcolor{GREEN}{$\uparrow$} & 24.26 \textcolor{GREEN}{$\uparrow$} & 75.33 \textcolor{RED}{$\downarrow$} & 40.95 \textcolor{GREEN}{$\uparrow$} & 22.48 \textcolor{GREEN}{$\uparrow$} & \underline{44.90} \textcolor{GREEN}{$\uparrow$} \\
    \rowcolor{gray!10} {Qwen3-8B} & Reasoning & 27.78 & 38.38 & 29.52 & 40.70 & 16.43 & 54.00 & 34.29 & 20.38 & 32.68 \\
     & FTRL-Reinforce++ & 38.11 \textcolor{GREEN}{$\uparrow$} & 43.70 \textcolor{GREEN}{$\uparrow$} & 39.40 \textcolor{GREEN}{$\uparrow$} & 41.24 \textcolor{GREEN}{$\uparrow$} & 32.52 \textcolor{GREEN}{$\uparrow$} & 53.52 \textcolor{RED}{$\downarrow$} & 37.52 \textcolor{GREEN}{$\uparrow$} & 22.29 \textcolor{GREEN}{$\uparrow$} & 38.54 \textcolor{GREEN}{$\uparrow$} \\
     & FTRL-GRPO & 40.67 \textcolor{GREEN}{$\uparrow$} & 40.06 \textcolor{GREEN}{$\uparrow$} & 38.58 \textcolor{GREEN}{$\uparrow$} & 40.50 \textcolor{RED}{$\downarrow$} & 28.91 \textcolor{GREEN}{$\uparrow$} & 54.57 \textcolor{GREEN}{$\uparrow$} & 39.24 \textcolor{GREEN}{$\uparrow$} & 22.95 \textcolor{GREEN}{$\uparrow$} & 38.19 \textcolor{GREEN}{$\uparrow$} \\
    \rowcolor{gray!10} {Qwen3-14B} & Reasoning & 31.91 & 40.23 & 32.96 & 40.10 & 18.87 & 57.33 & 36.19 & 20.29 & 34.74 \\
     & FTRL-Reinforce++ & 44.15 \textcolor{GREEN}{$\uparrow$} & 48.11 \textcolor{GREEN}{$\uparrow$} & 44.04 \textcolor{GREEN}{$\uparrow$} & 44.79 \textcolor{GREEN}{$\uparrow$} & 27.09 \textcolor{GREEN}{$\uparrow$} & 58.95 \textcolor{GREEN}{$\uparrow$} & 37.24 \textcolor{GREEN}{$\uparrow$} & 21.62 \textcolor{GREEN}{$\uparrow$} & 40.75 \textcolor{GREEN}{$\uparrow$} \\
     & FTRL-GRPO & 41.67 \textcolor{GREEN}{$\uparrow$} & 47.67 \textcolor{GREEN}{$\uparrow$} & 43.36 \textcolor{GREEN}{$\uparrow$} & 41.64 \textcolor{GREEN}{$\uparrow$} & 31.70 \textcolor{GREEN}{$\uparrow$} & 60.57 \textcolor{GREEN}{$\uparrow$} & 38.10 \textcolor{GREEN}{$\uparrow$} & 21.43 \textcolor{GREEN}{$\uparrow$} & 40.77 \textcolor{GREEN}{$\uparrow$} \\
    \bottomrule
    \end{tabular}
    }
    \caption{Performance of different LLMs on each test set. The best result for each dataset is \textbf{bolded}, and the second-best is \underline{underlined}. Performance improvements over the base model after training with our method (i.e., FTRL-Reinforce++ and FTRL-GRPO) are indicated with \textcolor{GREEN}{$\uparrow$}, while performance declines are marked with \textcolor{RED}{$\downarrow$}.}
    \label{tab:result}
\end{table*}

\paragraph{Complexity Scaling}
While integrating functionality leads to a more organized toolset, the resulting tools often exhibit relatively simple functionality and limited parameter sets~\cite{ToolHop}, which can constrain the model’s ability to generalize to more complex tools. To address this limitation, we enhance tool complexity through four key strategies:
1) Functional generalization: Expand each tool's function $f_i$ beyond its original sub-questions to cover a broader range of tasks, thereby increasing its versatility and data processing capability;
2) Parameter expansion: Enrich the parameter set $\mathbb{P}_i$ to support the extended functionality;
3) Parameter type generalization: Replace simple data types (e.g., strings) with more complex structures such as arrays, dictionaries, and nested types;
and 4) Toolset extension: Introduce additional non-essential tools to increase the complexity of tool selection and elevate the overall difficulty of the environment.
These enhancements result in a more realistic and challenging setting for tool use, better suited for training models capable of robust generalization.

\paragraph{Localized Deployment}
Once the tool document $d_i$ is finalized, it is mapped to a corresponding Python function and deployed locally. Key elements such as $n_i$, $pn_{ij}$, and $pt_{ij}$ are mapped to function name, arguments, and type annotations, respectively. Additionally, the associated $q_i$ and $a_i$ are used as prior conditions to ensure that the function returns correct outputs when invoked with valid arguments. This design establishes a stable and controlled environment, facilitating the effective learning of tool use through interaction.

Through these five stages, we are able to construct complete tool-use training environments without relying on external toolsets, significantly enhancing the scalability of training data. Since all tools are locally deployable, the environment can provide stable and consistent feedback. Moreover, by analyzing this feedback, we can accurately evaluate each step of the tool invocation process.

\subsection{Feedback-Driven Model Training}
\label{sec:train}

Building on the constructed training environments, we propose a feedback-driven training framework. By designing verifiable reward mechanisms that rely solely on environment feedback, it support various preference-based training strategies. 
This allows the model to continually improve its tool-use capabilities through interaction.

\paragraph{Verifiable Reward Design}
To enable continuous improvement, accurate and informative reward signals are essential~\cite{GRM}. Leveraging stable environments we construct, the model receives direct feedback after each tool invocation, which allows us to verify whether a sub-question has been successfully solved. This enables the generation of step-level, verifiable rewards.
To encourage both correctness and efficiency, we draw inspiration from the F1 score~\cite{f1}, balancing the precision of tool invocation and the completeness of task execution. We also assess the validity of the model’s final answer, resulting in a comprehensive reward signal. Importantly, the framework operates without external models or predefined solution paths, enhancing generality and applicability.
Specially, let $o$ be the model's output, $p$ be the number of tool invocations, $q \leq p$ be the number of sub-questions successfully solved, $t \geq 0$ be the number of remaining unsolved sub-questions, and $a$ be the correct final answer in the environment. The reward $R$ is defined as:
\[
R =
\begin{cases}
\frac{2q}{p + 1}, & \text{if } p > 0 \\
-0.5, & \text{elif } o \text{ is None} \\
-0.3, & \text{elif there is a format error} \\
\frac{1}{t + 1}, & \text{elif } a \text{ is in } o \\
0.5, & \text{elif } t = 0 \\
0, & \text{else}
\end{cases}
\]

\paragraph{Trajectory Data Collection}
Suppose we have an LLM $\mathcal{M}$ to be optimized, leveraging the reward mechanism we have designed, we use $\mathcal{M}$ to sample training trajectories. Specifically, $\mathcal{M}$ performs multi-step interactions with the constructed environment. At each step, we record the sampled trajectory, available tools, final answer, the remaining unsolved sub-questions, and their corresponding answers. This collection of information constitutes a single training instance used for optimization of $\mathcal{M}$'s behavior.

\paragraph{Preference-Based Training}
With the collected data and defined reward signals, we apply preference-based RL algorithms~\cite{PPO, GRPO, Reinforce++} to optimize the model’s tool-use policy, which encourage behavior that improves along reward gradients.
Through repeated interaction and feedback, the model can progressively improve its precision in tool invocation, task-solving ability, and final output validity—without the need for manually annotated solution paths.

\section{Experimental Setup}
In this section, we present our experimental setup.\footnote{Additional details are available in Appendix~\ref{sec:detail-setup}.}

\paragraph{Datasets}
We train the model on our self-constructed dataset and evaluate it across four distinct tool-use benchmarks. These include \textbf{Ours} as the in-domain test set, and \textbf{ToolHop}~\cite{ToolHop}, \textbf{$\tau$-bench}~\cite{tau-bench}, and \textbf{RoTBench}~\cite{RoTBench} as out-of-domain test sets. Detailed dataset information is provided in Table~\ref{tab:datasets}.

\begin{table}[!t]
    \centering
    \resizebox{\linewidth}{!}
    {
    \begin{tabular}{ccccc}
    \toprule
    \textbf{Layer} & \textbf{Module} & \textbf{Sub-Module} & \textbf{Count} & \textbf{Cumulation} \\ \midrule
    1 & MLP & down\_proj.weight & 11& 30.56 \\
    2 & MLP & up\_proj.weight & 8 & 52.78 \\
    2 & MLP & down\_proj.weight & 6 & 69.44 \\
    1 & MLP & up\_proj.weight & 3 & 77.78 \\
    0 & MLP & down\_proj.weight & 2 & 83.33 \\
    26 & Self\_Attn & k\_proj.weight & 2 & 88.89 \\
    3 & MLP & down\_proj.weight & 1 & 91.67 \\
    9 & Self\_Attn & v\_proj.bias & 1 & 94.44 \\
    13 & Self\_Attn & v\_proj.bias & 1 & 97.22 \\
    17 & Self\_Attn & v\_proj.bias & 1 & 100.00 \\
    \bottomrule
    \end{tabular}
    }
    \caption{Distribution of the top three modules with the highest relative parameter change rates before and after training, ordered by descending proportion.}
    \label{tab:parameter}
\end{table}

\paragraph{Metrics}
We adopt the original evaluation metrics defined for each dataset to objectively assess the performance of LLMs. 
For \emph{Ours}, we use \textbf{Solve-P} to measure the precision of tool invocations, \textbf{Solve-R} to evaluate task completeness, and \textbf{Solve-F1}, the harmonic mean of Solve-P and Solve-R, to provide an overall performance score. 
For \emph{ToolHop}, we use \textbf{Answer Correctness (AC)} to assess the model’s ability to correctly use tools in answering multi-hop questions. 
For \emph{$\tau$-bench}, we report \textbf{Pass\texttt{\char`^}1}, which estimates the probability that the model produces the correct answer. 
For \emph{RoTBench}, we use three metrics: \textbf{Tool Selection (TS)} to evaluate whether the model selects the correct tool, \textbf{Parameter Identification (PI)} to assess whether the correct parameters are identified given the correct tool, and \textbf{Content Filling (CF)} to measure the accuracy of parameter value generation, conditional on correct tool and parameter selection.

\paragraph{Baselines}
To enable a comprehensive comparison of model performance, we evaluate 12 representative LLMs. From \emph{closed-source LLMs}, we include \textbf{Gemini-2.5-Flash}~\cite{Gemini2.5}, \textbf{Gemini-2.5-Pro}~\cite{Gemini2.5}, \textbf{Claude-4.0-Sonnet}~\cite{Claude4.0}, \textbf{GPT-3.5-Turbo}~\cite{GPT-3}, and \textbf{GPT-4o}~\cite{GPT-4}. From \emph{open-source LLMs}, we include the \textbf{Qwen2.5}~\cite{Qwen2.5} and \textbf{Qwen3}~\cite{Qwen3} series, covering both \textbf{non-reasoning} and \textbf{reasoning} modes. Additionally, we include two versions of \emph{our own approaches}: \textbf{FTRL-Reinforce++} and \textbf{FTRL-GRPO}, which apply the Reinforce++~\cite{Reinforce++} and GRPO~\cite{GRPO} algorithms, respectively, within our training framework.

\begin{table}[!t]
\centering
\resizebox{0.85\linewidth}{!}
{
\begin{tabular}{lcc}
\toprule
\textbf{Dataset}  &  \textbf{Capability}   &  \textbf{\# Number} \\ \midrule
MMLU~\cite{MMLU} & General & 14042 \\
BBH~\cite{BBH} & General & 6511 \\
GSM8K~\cite{GSM8K} & Reasoning & 1319 \\
MATH~\cite{MATH} & Reasoning & 5000 \\
HumanEval~\cite{HumanEval} & Coding & 164 \\
MBPP~\cite{MBPP} & Coding & 257 \\
\bottomrule
\end{tabular}
}
\caption{Datasets used for testing general performance.}
\label{tab:general}
\end{table}

\begin{figure*}[!t]
    \centering
    \includegraphics[width=\linewidth]{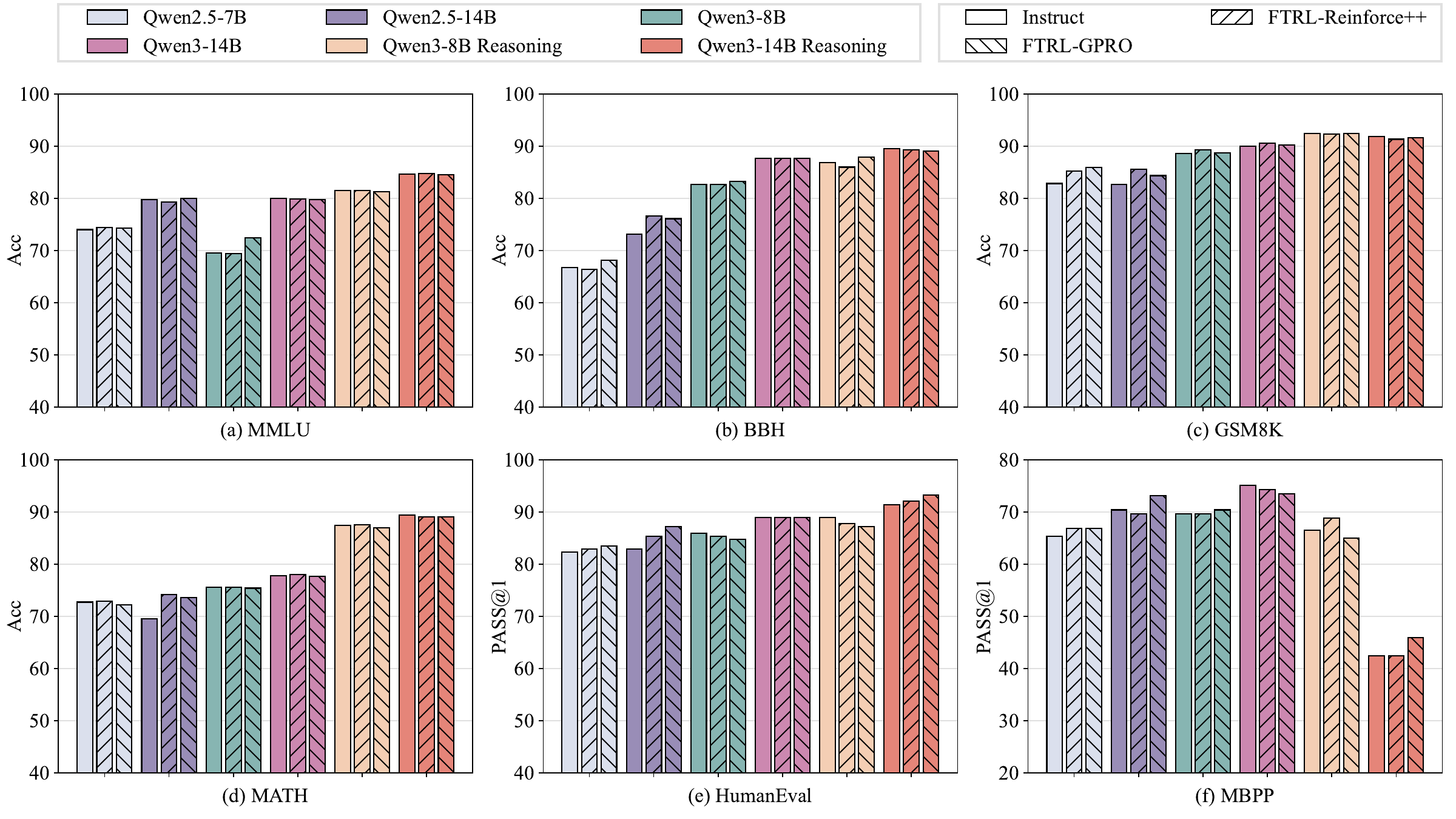}
    \caption{Performance of each generalized capability before and after training across different models.}
    \label{fig:general}
\end{figure*}

\paragraph{Implementation details}
In the \textbf{environment construction} stage, we manually create user inputs and use GPT-4o to assist in building the environments.\footnote{Detailed prompts are provided in Appendix~\ref{sec:prompt-env}.} To ensure stable and reproducible results, we set the temperature to 0 and the max\_tokens parameter to 2048.
In the \textbf{training} stage, we employ the VeRL~\cite{VeRL} framework with the following hyperparameters: a learning rate of $1\mathrm{e}{-6}$, batch size of 512, mini-batch size of 32, and 16 rollouts per update. The maximum response length per step is set to 1024 for non-reasoning mode and 8192 for reasoning mode. Training is performed over three epochs, with training trajectories resampled at the beginning of each epoch using the current model. All training is conducted on 8 NVIDIA A100 GPUs. Due to resource constraints, we restrict all open-source LLMs used in training to the 7B–14B parameter range.
In the \textbf{test} stage, we ensure consistent and stable results by using each model’s built-in chat template,
applying greedy decoding, and setting the maximum response length per step to 1024 for non-reasoning mode and 8192 for reasoning mode. For evaluation on $\tau$-bench, we simulate the user using GPT-4o, with temperature set to 0 and max\_tokens set to 512.

\section{Main Results}
\label{sec:result}

As shown in Table~\ref{tab:result}, we evaluate the performance of various LLMs.\footnote{Detailed performance by scenario is provided in Appendix~\ref{sec:detail-result}.} Based on it, we make the following observations.

\textbf{Our approach consistently enhances the model's tool-use capabilities across various conditions.}
Across different model families, RL algorithms, and inference modes, our method reliably improves tool-use performance and demonstrates strong generalization on out-of-domain data. On average, it yields over a 10\% performance gain to LLMs across multiple benchmarks. Remarkably, while open-source LLMs generally underperform compared to closed-source LLMs, models with 8B and 14B parameters trained using our method outperform the strongest available closed-source models on average. Furthermore, despite the absence of multi-turn user interactions and noisy environments in our training data, the models still achieved substantial performance gains on both $\tau$-bench and RoTBench. These results suggest that our approach effectively enhances the model’s ability to use tools, even under challenging or previously unseen conditions.

\textbf{Performance gains achieved by our method appear to primarily stem from updates to the model’s lower-layer MLP parameters.}
To better understand why our approach enhances tool-use capabilities, we conduct a parameter-level analysis. Following~\citet{MulDimIF}, we compare the relative update rates of different parameter modules before and after training across 12 trained LLMs. For each model, we identify the top three modules with the highest rates of change and summarize their distribution in Table~\ref{tab:parameter}. Interestingly, we find that most of these frequently updated modules are concentrated in the MLP components of the lower layers (i.e., layers 0–2). This suggests that our method improves performance primarily by enhancing the model's ability to understand and represent contextual information in the early stages of processing.\footnote{Case studies are provided in Appendix~\ref{sec:case-study}.}

\textbf{Current open-source LLMs do not necessarily exhibit stronger tool-use performance in reasoning mode compared to non-reasoning mode.}
Although reasoning mode has demonstrated effectiveness in enhancing the performance of Qwen3-family models on complex tasks~\cite{Qwen3}, it does not consistently offer advantages for tool use. Specifically, while reasoning mode improves performance on datasets such as ToolHop and $\tau$-bench, it also leads to a notable reduction in RoTBench. An in-depth analysis of our dataset reveals that reasoning mode boosts performance in multi-hop and parallel multi-hop scenarios but significantly degrades performance in the single-hop case.\footnote{Details can be found in Table~\ref{tab:self-result} in Appendix~\ref{sec:detail-result}.} This suggests that the current implementation of reasoning enhances the model's capability in complex scenarios at the cost of performance in simpler ones. This trade-off likely stems from the fact that existing reasoning mechanisms in open-source LLMs are primarily optimized for mathematically tasks, limiting their adaptability across diverse tool-use settings. These findings underscore a gap in current modeling strategies and highlight the need for more refined reasoning mechanisms for tool use.

\begin{figure}
    \centering
    \includegraphics[width=\linewidth]{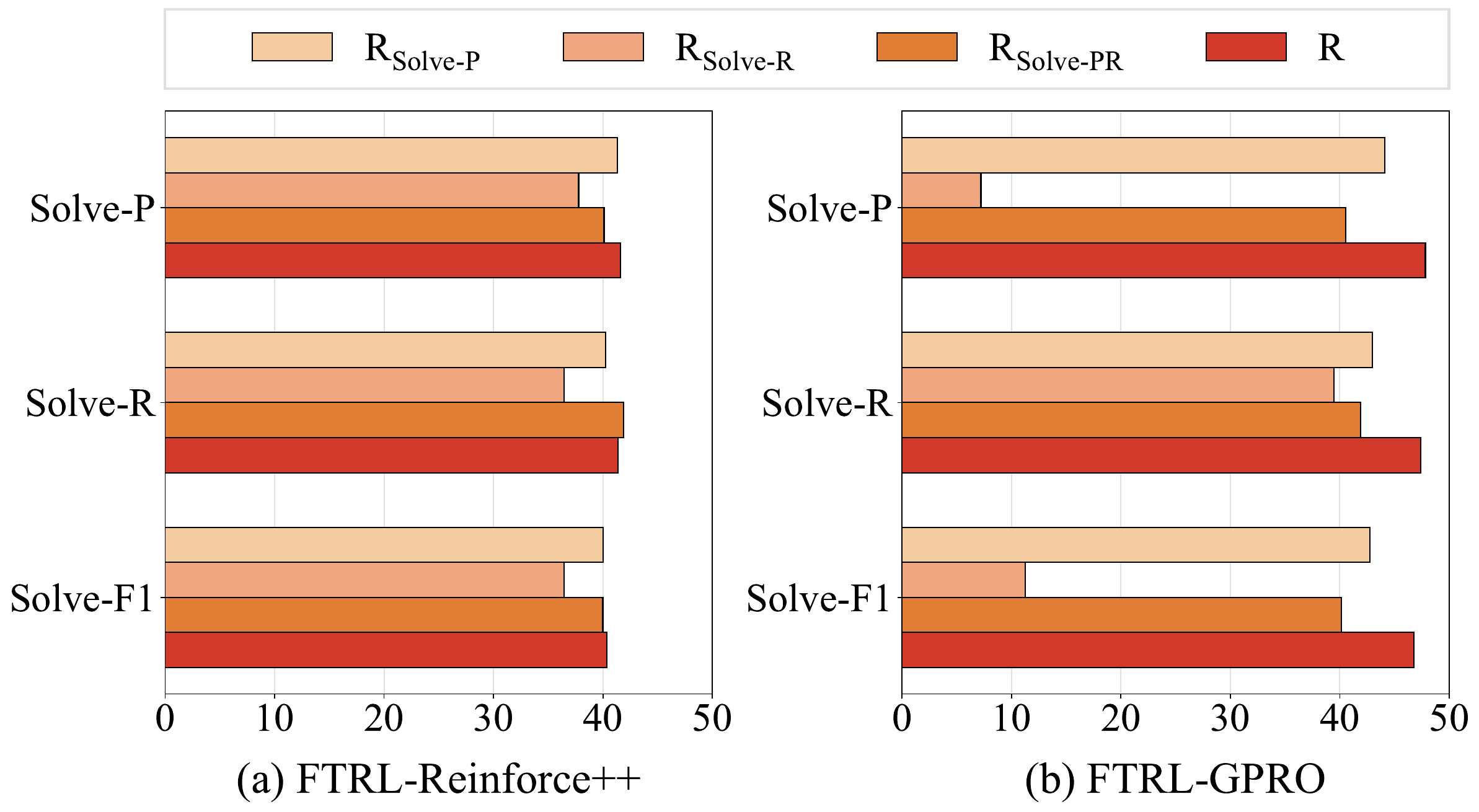}
    \caption{Performance of Qwen 2.5-7B trained using different reward mechanisms.}
    \label{fig:reward}
\end{figure}

\section{Further Studies}

In this section, we compare the general capabilities of the models before and after training. Additionally, we evaluate the impact of different reward mechanisms on performance, providing empirical justification for our chosen design. Furthermore, we analyze performance trends throughout the training process.

\paragraph{Performance on General Tasks}
To evaluate the impact of our approach on general abilities alongside its improvements in tool-use performance, we evaluate models before and after training on the six public test sets listed in Table~\ref{tab:general}. As shown in Figure~\ref{fig:general}, training with our method significantly enhances tool-use capabilities without compromising generalization performance. These results suggest that our method improves the model's ability to understand and represent contextual information, rather than merely overfitting to the training data. This indicates that our approach is compatible with existing LLM training paradigms and holds broad potential for real-world applications.

\paragraph{Impact of Reward Mechanisms}
The design of the reward mechanism is a critical component of any RL algorithm, as it directly influences the performance of the trained model~\cite{R1, GRM}. As described in Section~\ref{sec:train}, our reward mechanism is designed to balance both the precision of tool invocations and overall task completion. To validate the effectiveness of this design, we compare the performance of Qwen2.5-7B trained with four distinct reward functions, each differing only in how tool-calling behavior is scored:
1) $R_{\text{Solve-P}} = \frac{q}{p}$: focuses solely on precision;
2) $R_{\text{Solve-R}} = q$: rewards only for task completion;
3) $R_{\text{Solve-PR}} = \frac{q \cdot q}{p}$: directly multiplies task completion with precision;
and 4) $R = \frac{2q}{p+1}$: our proposed reward, which balances both precision and completeness.
As shown in Figure~\ref{fig:reward}, optimizing for precision alone (i.e., $R_{\text{Solve-P}}$) leads to incomplete task execution, while optimizing for completion alone (i.e., $R_{\text{Solve-R}}$) can severely degrade tool precision, as the model tends to overuse tools to maximize reward. Although $R_{\text{Solve-PR}}$ incorporates both components, its discrete reward distribution hinders stable training. In contrast, our proposed reward function enables a better balance between tool-use precision and task completion, resulting in improved overall performance.

\begin{figure}
    \centering
    \includegraphics[width=\linewidth]{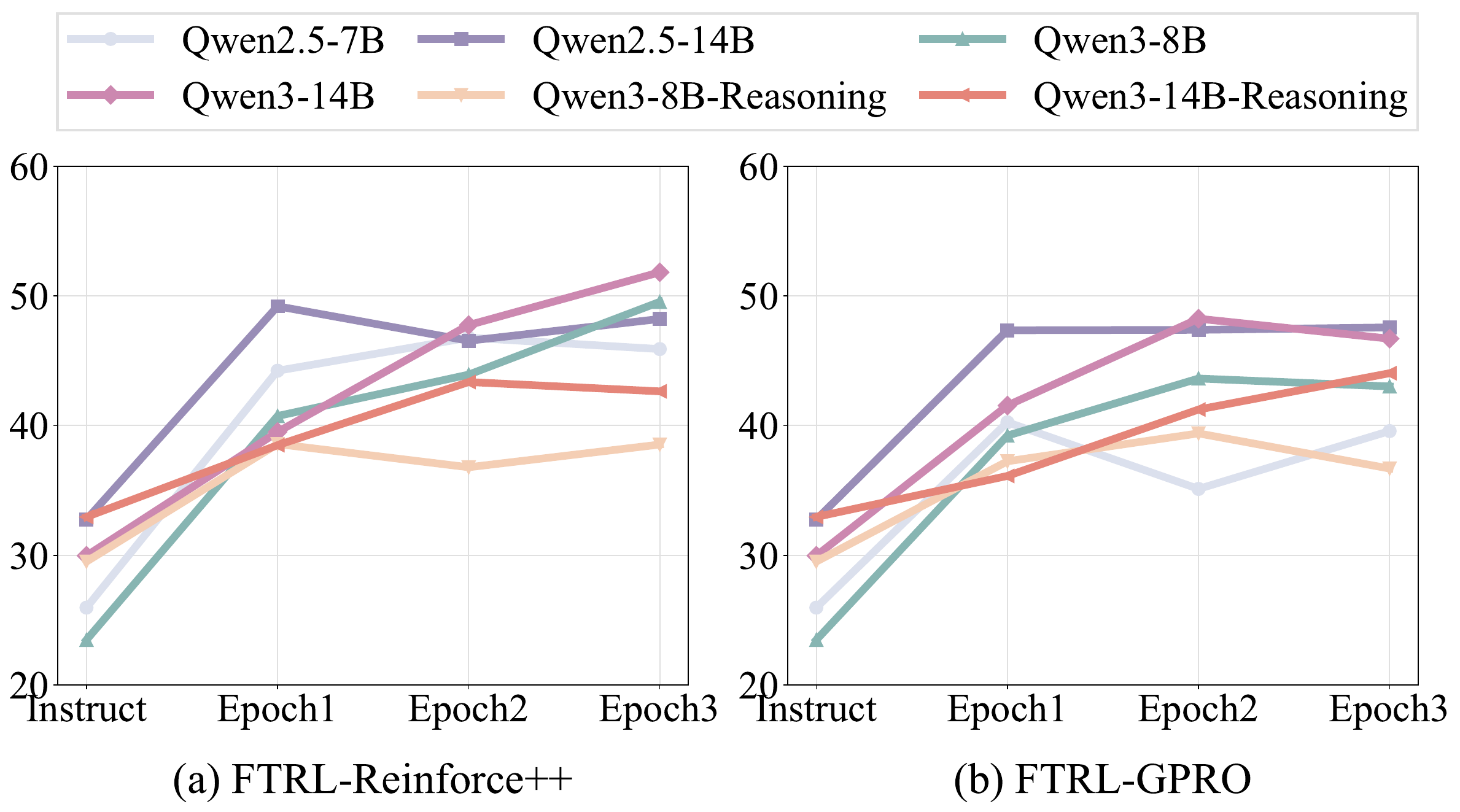}
    \caption{Solve-F1 of various LLMs.}
    \label{fig:iter}
\end{figure}

\paragraph{Effect of Iteration Count}
We train each model over three epochs, resampling new training trajectories at the beginning of each epoch to expand the exploration space. Figure~\ref{fig:iter} presents model performance across training epochs. The results show substantial performance improvements after the first epoch, highlighting the efficiency of our data and training method. Furthermore, as training progresses, most models continue to exhibit consistent performance gains, indicating that our strategy maintains a sufficiently rich exploration space. This leads to better data utilization and enhanced training efficiency.

\section{Conclusion}

In this paper, we propose an automated strategy for constructing tool-use training environments that progresses through a five-stage pipeline. This approach enables the creation of diverse and comprehensive training settings without relying on external toolsets. Building on these environments, we introduce a feedback-driven training framework that enhances a model’s tool-use capabilities by leveraging a verifiable reward function. This reward balances tool invocation precision and task completion, and relies solely on feedback from the environment. Extensive experimental results demonstrate the effectiveness and generalizability of our method. Furthermore, analysis suggests that the observed performance gains primarily stem from updates to the model’s lower-layer MLP parameters, suggesting improved contextual understanding during tool interaction.

\section*{Limitations}

Although we propose an automated pipeline for constructing tool-use training environments and a feedback-driven training framework that effectively enhance a model's tool-use capabilities, our approach primarily focuses on improving tool invocation rather than optimizing the model's underlying reasoning process. As discussed in Section~\ref{sec:result}, the reasoning patterns of current open-source models are not well aligned with the tool use task, resulting in a significant gap between their reasoning behavior and actual tool-use performance. Therefore, it is a key direction for us to explore methods to better align and optimize the reasoning process for tool use.

\section*{Acknowledgments}

The authors wish to thank the anonymous reviewers for their helpful comments. This work was partially funded by Henan Province Major Industrial ``Challenge-Based Innovation'' (No. 251000210300), National Natural Science Foundation of China (No. 62476061, 62376061, 62576106).


\bibliographystyle{plainnat}
\bibliography{main}


\beginappendix

\section{Details of Experimental Setup}
\label{sec:detail-setup}

In this section, we detail our experimental setup, including the datasets used (Section~\ref{sec:datasets}), metrics for each benchmark (Section~\ref{sec:metrics}), and baselines for comparison (Section~\ref{sec:baselines}).

\subsection{Datasets}
\label{sec:datasets}

\begin{table*}[!t]
    \centering
    \resizebox{\linewidth}{!}
    {
    \begin{tabular}{l cccc cccc}
    \toprule
    \multirow{2}*{\textbf{Environments}}     & \multicolumn{4}{c}{\textbf{Train}} & \multicolumn{4}{c}{\textbf{Test}}  \\ \cmidrule(lr){2-5} \cmidrule(lr){6-9}
         & \textbf{Single-Hop} & \textbf{Parallel Single-Hop} & \textbf{Multi-Hop} & \textbf{Parallel Multi-Hop} & \textbf{Single-Hop} & \textbf{Parallel Single-Hop} & \textbf{Multi-Hop} & \textbf{Parallel Multi-Hop} \\ \midrule
    \# Questions & 500	&500	&500	&715	&50	&50	&50	&50\\
    \# Avg Sub-Qs & 1.00 	&2.10 	&4.71 	&6.97 	&1.00 	&2.02 	&5.72 	&7.66 \\
    \# Avg Tools & 8.00 	&8.14 	&9.81 	&10.92 	&7.96 	&7.48 	&10.34 	&11.26 \\
    \bottomrule
    \end{tabular}
    }
    \caption{Statistical overview of constructed environments. `\# Questions' indicates the total number of questions; `\# Avg Sub-Qs' refers to the average number of sub-questions derived from each question; and `\# Avg Tools' denotes the average number of tools associated with each data instance.}
    \label{tab:ours}
\end{table*}

\paragraph{Train}
Based on the automated environment construction scheme described in Section~\ref{sec:env}, we create diverse training data across four distinct scenarios, each paired with a corresponding locally executable toolset. This results in a comprehensive and varied set of training environments. Detailed data distributions are provided in Table~\ref{tab:ours}.

\paragraph{Test}
To comprehensively evaluate the performance gains introduced by our approach, we construct a custom test set for in-domain evaluation and employ three publicly available tool-use datasets for out-of-domain evaluation.
\begin{itemize}
    \item \textbf{Ours} refers to the dataset constructed using the scheme described in Section~\ref{sec:env}, containing 50 entries per scenario. This dataset serves as the in-domain test set for evaluating model performance. We ensure that there is no overlap between the test and training sets.
    \item \textbf{ToolHop} is the first dataset specifically designed to evaluate model performance in the multi-hop tool use task. It contains 995 data instances and defines three scenarios to comprehensively assess a model's ability to use tools to answer multi-hop questions under varying conditional constraints.
    \item \textbf{$\tau$-bench} is a dataset designed to evaluate a model’s ability to use various data manipulation tools to meet user needs related to airline ticketing or book reservations. It is characterized by multi-turn interactions between the user and the model and contains a total of 165 data instances.
    \item \textbf{RoTBench} is a dataset designed to evaluate the robustness of LLMs in tool use. It includes five different noise-level scenarios and consists of a total of 840 data instances.
\end{itemize}

\subsection{Metrics}
\label{sec:metrics}
We adopt the original evaluation metrics defined for each dataset to objectively evaluate the performance of LLMs.

\paragraph{Ours}
Assume that for a given test instance, there are $n$ sub-questions in total. During its interaction with the environment, the model invokes tools $p$ times and successfully solves $q \leq p$ sub-questions. We evaluate the model’s performance using the following three metrics:

\begin{itemize}
  \item \textbf{Solve-P}: 
  \[
  \text{Solve-P} = 
  \begin{cases}
    \frac{q}{p}, & \text{if } p > 0 \\
    1, & \text{if } p = 0
  \end{cases}
  \]
  This metric measures the precision of the tool invocations.
  
  \item \textbf{Solve-R}:
  \[
  \text{Solve-R} = \frac{q}{n}
  \]
  This metric measures the completeness of the task.

  \item \textbf{Solve-F1}:
  \[
  \text{Solve-F1} = \frac{2 \cdot \text{Solve-P} \cdot \text{Solve-R}}{\text{Solve-P} + \text{Solve-R}}
  \]
  This computes the harmonic mean of Solve-P and Solve-R, providing an overall performance measure.
\end{itemize}

\paragraph{ToolHop}
Assume that for a test instance, the standard answer is $a$ and the model's final response is $o$. We evaluate the model's performance using the following metric:

\begin{itemize}
  \item \textbf{Answer Correctness (AC)}:
  \[
  \text{AC} = 
  \begin{cases}
    1, & \text{if } a \text{ is in } o \\
    0, & \text{otherwise}
  \end{cases}
  \]
  This metric assesses the model's ability to correctly use tools to answer multi-hop questions.
\end{itemize}

\paragraph{$\tau$-bench}
Assume that for a test set, the model makes a total of $num$ attempts, out of which $c$ are correct. We evaluate the model's performance using the following metric:

\begin{itemize}
  \item \textbf{Pass\texttt{\char`^}1}:
  \[
  \text{Pass\^{}1} = \frac{c}{num}
  \]
  This metric estimates the probability that the model produces the correct answer on its first attempt.
\end{itemize}

\paragraph{RoTBench}
Assume that for a test instance, the correct tool is $t$, the correct parameter set is $\mathbb{P}$, and the corresponding parameter values are $\mathbb{V}$. The model predicts a tool $t'$, a parameter set $\mathbb{P'}$, and parameter values $\mathbb{V'}$. We evaluate the model's performance using the following three metrics:

\begin{itemize}
  \item \textbf{Tool Selection (TS)}:
  \[
  \text{TS} =
  \begin{cases}
    1, & \text{if } t = t' \\
    0, & \text{otherwise}
  \end{cases}
  \]
  This metric measures the model's ability to select the correct tool.

  \item \textbf{Parameter Identification (PI)}:
  \[
  \text{PI} =
  \begin{cases}
    \text{TS}, & \text{if } \mathbb{P} = \mathbb{P'} \\
    0, & \text{otherwise}
  \end{cases}
  \]
  This metric evaluates whether the model correctly identifies the required parameter set, conditional on correct tool selection.

  \item \textbf{Content Filling (CF)}:
  \[
  \text{CF} =
  \begin{cases}
    \text{PI}, & \text{if } \mathbb{V} = \mathbb{V'} \\
    0, & \text{otherwise}
  \end{cases}
  \]
  This metric assesses the model's ability to accurately fill in the parameter values, conditional on correct tool and parameter set selection.
\end{itemize}

\subsection{Baselines}
\label{sec:baselines}

To enable a comprehensive comparison of model performance, we select 12 representative LLMs for evaluation.

\paragraph{Closed-Source LLMs}
We select five of the most representative closed-source models for evaluation, reflecting the current state-of-the-art in LLMs.

\begin{itemize}
  \item \textbf{Gemini-2.5-Flash} and \textbf{Gemini-2.5-Pro}, the latest generation of LLMs released by Google, demonstrating strong performance across a wide range of complex tasks.
  \item \textbf{Claude-4.0-Sonnet}, developed by Anthropic, known for its exceptional capabilities in code-related tasks.
  \item \textbf{GPT-3.5-Turbo} and \textbf{GPT-4o}, released by OpenAI, which excel at a broad spectrum of general-purpose tasks.
\end{itemize}

\paragraph{Open-Source LLMs}
We select all models ranging from 7B to 72B in the \textbf{Qwen2.5} and \textbf{Qwen3} series as representative open-source models for evaluation, as they reflect the current state-of-the-art among open-source LLMs of similar scale. Additionally, since the Qwen3 models support both \textbf{reasoning} and \textbf{non-reasoning} modes, we evaluate them under both configurations to comprehensively evaluate their performance.

\paragraph{Ours}
To demonstrate the generalizability of our approach, we apply the Reinforce++ and GRPO algorithms to our training framework, resulting in \textbf{FTRL-Reinforce++} and \textbf{FTRL-GRPO}. Considering resource constraints, we conduct experiments using these algorithms on all open-source LLMs range in the 7B--14B.


\section{Detailed Results for Each Dataset}
\label{sec:detail-result}

We conduct a comprehensive and detailed evaluation of the tool-use capabilities of various LLMs on a self-constructed in-domain dataset and three publicly available out-of-domain datasets. The detailed results for each dataset are presented from Table~\ref{tab:self-result} to Table~\ref{tab:rotbench-result}.

\section{Case Study}
\label{sec:case-study}
To clearly demonstrate the effectiveness of our method in improving tool-use performance in LLMs, this section presents and analyzes several representative cases. Specifically, Table~\ref{tab:qwen2.5-7b-ftrl-grpo-vs-qwen2.5-7b} and Table~\ref{tab:qwen3-8b-reasoning-reinforce++-vs-qwen3-8b-reasoning} compare the performance of Qwen2.5-7B-Instruct with Qwen2.5-7B-FTRL-GRPO, and Qwen3-8B-Reasoning with Qwen3-8B-Reasoning-FTRL-Reinforce++, highlighting the advantages of our approach in enhancing contextual understanding and decision-making capabilities. Furthermore, Table~\ref{tab:qwen3-8b-vs-qwen3-8b-reasoning-event-location} compares Qwen3-8B’s performance in reasoning versus non-reasoning modes, reinforcing our claim from Section~\ref{sec:result} that existing reasoning modes are insufficient for effectively distinguishing between different tools.

\section{Prompts for Environment Construction}
\label{sec:prompt-env}

With the support of GPT-4o’s powerful natural language understanding capabilities, we successfully carry out automated environment construction. The prompts used in this process are listed from Figure~\ref{fig:prompt-dg} to Figure~\ref{fig:prompt-cv}.



\begin{table*}[h]
    \centering
    \resizebox{\linewidth}{!}{
    \begin{tabular}{ll lll lll lll lll lll}
    \toprule
    \multirow{2}*{\textbf{Family}} & \multirow{2}*{\textbf{Version}} & \multicolumn{3}{l}{\textbf{Single-Hop}} & \multicolumn{3}{l}{\textbf{Parallel-Single-Hop}} & \multicolumn{3}{l}{\textbf{Multi-Hop}} & \multicolumn{3}{l}{\textbf{Parallel-Multi-Hop}} & \multicolumn{3}{l}{\textbf{Avg}} \\ 
    \cmidrule(l){3-5} \cmidrule(l){6-8} \cmidrule(l){9-11} \cmidrule(l){12-14} \cmidrule(l){15-17}
     & & \textbf{Solve-P} & \textbf{Solve-R} & \textbf{Solve-F1} & \textbf{Solve-P} & \textbf{Solve-R} & \textbf{Solve-F1} & \textbf{Solve-P} & \textbf{Solve-R} & \textbf{Solve-F1} & \textbf{Solve-P} & \textbf{Solve-R} & \textbf{Solve-F1} & \textbf{Solve-P} & \textbf{Solve-R} & \textbf{Solve-F1} \\
    \midrule
    \multicolumn{17}{c}{\textit{Closed-Source LLMs}} \\
    Gemini & 2.5-Flash & 37.00 & 20.00 & 18.00 & \textbf{40.67} & 32.00 & 30.50 & \textbf{68.80} & 15.26 & 16.89 & 64.65 & 25.28 & 27.59 & \textbf{52.78} & 23.13 & 23.24 \\
    Gemini & 2.5-Pro & \textbf{46.67} & 30.00 & 27.67 & 37.22 & 41.00 & 32.81 & \underline{59.57} & 18.11 & 20.75 & 63.64 & 38.72 & 38.68 & \underline{51.77} & 31.96 & 29.98 \\
    Claude & 4.0-Sonnet & 25.17 & 36.00 & 26.73 & 35.14 & \textbf{53.00} & 35.75 & 54.33 & 56.02 & 45.09 & 54.67 & \textbf{79.24} & 62.77 & 42.33 & 56.07 & 42.59 \\
    GPT & 3.5-Turbo & 23.50 & 32.00 & 25.80 & 25.14 & 33.33 & 27.49 & 19.21 & 23.37 & 19.14 & 35.23 & 28.45 & 30.22 & 25.77 & 29.29 & 25.66 \\
    GPT & 4o & 18.83 & 26.00 & 19.47 & 28.44 & 40.33 & 30.48 & 35.34 & 43.24 & 31.48 & 51.09 & 58.46 & 52.13 & 33.43 & 42.01 & 33.39 \\
    \midrule
    \multicolumn{17}{c}{\textit{Open-Source LLMs (32B - 72B)}} \\
    Qwen2.5-32B & Instruct & 21.29 & 26.00 & 22.50 & 33.60 & 43.33 & 36.85 & 25.55 & 33.55 & 28.03 & 40.06 & 48.32 & 42.29 & 30.12 & 37.80 & 32.42 \\
    Qwen2.5-72B & Instruct & 27.07 & 34.00 & 29.00 & 24.17 & 36.67 & 27.37 & 32.51 & 45.72 & 36.52 & 42.98 & 57.26 & 47.97 & 31.68 & 43.41 & 35.22 \\
    Qwen3-32B & Non-Reasoning & 29.97 & 44.00 & 33.31 & 25.39 & 46.00 & 30.09 & 33.43 & 42.36 & 35.95 & 44.01 & 52.60 & 46.00 & 33.20 & 46.24 & 36.34 \\
    Qwen3-32B & Reasoning & 18.33 & 28.00 & 20.93 & 22.36 & 40.00 & 25.81 & 30.87 & 39.11 & 32.28 & 44.20 & 54.53 & 46.19 & 28.94 & 40.41 & 31.30 \\
    \midrule
    \multicolumn{17}{c}{\textit{Open-Source LLMs (7B - 14B)}} \\
    \rowcolor{gray!10} Qwen2.5-7B & Instruct & 19.83 & 26.00 & 21.47 & 27.30 & 29.67 & 26.81 & 25.59 & 25.46 & 24.09 & 37.03 & 29.63 & 31.52 & 27.44 & 27.69 & 25.97 \\
     & FTRL-Reinforce++ & 27.67 \textcolor{GREEN}{$\uparrow$} & 30.00 \textcolor{GREEN}{$\uparrow$} & 28.33 \textcolor{GREEN}{$\uparrow$} & 31.37 \textcolor{GREEN}{$\uparrow$} & 34.33 \textcolor{GREEN}{$\uparrow$} & 31.93 \textcolor{GREEN}{$\uparrow$} & 44.74 \textcolor{GREEN}{$\uparrow$} & 43.08 \textcolor{GREEN}{$\uparrow$} & 42.03 \textcolor{GREEN}{$\uparrow$} & 62.66 \textcolor{GREEN}{$\uparrow$} & 58.04 \textcolor{GREEN}{$\uparrow$} & 59.16 \textcolor{GREEN}{$\uparrow$} & 41.61 \textcolor{GREEN}{$\uparrow$} & 41.36 \textcolor{GREEN}{$\uparrow$} & 40.36 \textcolor{GREEN}{$\uparrow$} \\
     & FTRL-GRPO & 41.00 \textcolor{GREEN}{$\uparrow$} & 42.00 \textcolor{GREEN}{$\uparrow$} & 40.00 \textcolor{GREEN}{$\uparrow$} & 37.67 \textcolor{GREEN}{$\uparrow$} & 40.00 \textcolor{GREEN}{$\uparrow$} & 38.53 \textcolor{GREEN}{$\uparrow$} & 47.26 \textcolor{GREEN}{$\uparrow$} & 47.76 \textcolor{GREEN}{$\uparrow$} & 47.11 \textcolor{GREEN}{$\uparrow$} & 65.35 \textcolor{GREEN}{$\uparrow$} & 59.71 \textcolor{GREEN}{$\uparrow$} & 61.47 \textcolor{GREEN}{$\uparrow$} & 47.82 \textcolor{GREEN}{$\uparrow$} & 47.37 \textcolor{GREEN}{$\uparrow$} & 46.78 \textcolor{GREEN}{$\uparrow$} \\
    \rowcolor{gray!10} Qwen2.5-14B & Instruct & 25.50 & 26.00 & 24.13 & 24.85 & 30.67 & 26.71 & 32.58 & 33.31 & 31.86 & 50.30 & 48.57 & 48.33 & 33.31 & 34.64 & 32.76 \\
     & FTRL-Reinforce++ & 34.00 \textcolor{GREEN}{$\uparrow$} & 36.00 \textcolor{GREEN}{$\uparrow$} & 34.67 \textcolor{GREEN}{$\uparrow$} & 34.67 \textcolor{GREEN}{$\uparrow$} & 38.33 \textcolor{GREEN}{$\uparrow$} & 36.13 \textcolor{GREEN}{$\uparrow$} & 55.20 \textcolor{GREEN}{$\uparrow$} & 48.59 \textcolor{GREEN}{$\uparrow$} & 50.42 \textcolor{GREEN}{$\uparrow$} & \textbf{74.16} \textcolor{GREEN}{$\uparrow$} & 65.86 \textcolor{GREEN}{$\uparrow$} & \textbf{69.07} \textcolor{GREEN}{$\uparrow$} & 49.51 \textcolor{GREEN}{$\uparrow$} & 47.20 \textcolor{GREEN}{$\uparrow$} & 47.57 \textcolor{GREEN}{$\uparrow$} \\
     & FTRL-GRPO & 36.00 \textcolor{GREEN}{$\uparrow$} & 38.00 \textcolor{GREEN}{$\uparrow$} & 36.67 \textcolor{GREEN}{$\uparrow$} & \underline{40.00} \textcolor{GREEN}{$\uparrow$} & 41.33 \textcolor{GREEN}{$\uparrow$} & \underline{40.47} \textcolor{GREEN}{$\uparrow$} & 52.07 \textcolor{GREEN}{$\uparrow$} & 56.39 \textcolor{GREEN}{$\uparrow$} & 53.53 \textcolor{GREEN}{$\uparrow$} & 66.62 \textcolor{GREEN}{$\uparrow$} & 66.53 \textcolor{GREEN}{$\uparrow$} & 66.13 \textcolor{GREEN}{$\uparrow$} & 48.67 \textcolor{GREEN}{$\uparrow$} & 50.56 \textcolor{GREEN}{$\uparrow$} & 49.20 \textcolor{GREEN}{$\uparrow$} \\
    \rowcolor{gray!10} Qwen3-8B & Non-Reasoning & 31.15 & 40.00 & 33.29 & 26.58 & 37.67 & 28.69 & 8.86 & 16.12 & 10.56 & 18.12 & 29.05 & 21.37 & 21.18 & 30.71 & 23.48 \\
     & FTRL-Reinforce++ & 33.00 \textcolor{GREEN}{$\uparrow$} & 40.00 & 35.00 \textcolor{GREEN}{$\uparrow$} & 35.73 \textcolor{GREEN}{$\uparrow$} & 40.33 \textcolor{GREEN}{$\uparrow$} & 37.23 \textcolor{GREEN}{$\uparrow$} & 39.28 \textcolor{GREEN}{$\uparrow$} & 45.22 \textcolor{GREEN}{$\uparrow$} & 40.94 \textcolor{GREEN}{$\uparrow$} & 59.81 \textcolor{GREEN}{$\uparrow$} & 63.94 \textcolor{GREEN}{$\uparrow$} & 61.35 \textcolor{GREEN}{$\uparrow$} & 41.96 \textcolor{GREEN}{$\uparrow$} & 47.37 \textcolor{GREEN}{$\uparrow$} & 43.63 \textcolor{GREEN}{$\uparrow$} \\
     & FTRL-GRPO & \underline{43.00} \textcolor{GREEN}{$\uparrow$} & 50.00 \textcolor{GREEN}{$\uparrow$} & \underline{45.00} \textcolor{GREEN}{$\uparrow$} & 35.92 \textcolor{GREEN}{$\uparrow$} & 46.33 \textcolor{GREEN}{$\uparrow$} & 39.31 \textcolor{GREEN}{$\uparrow$} & 50.15 \textcolor{GREEN}{$\uparrow$} & \underline{60.21} \textcolor{GREEN}{$\uparrow$} & \underline{53.61} \textcolor{GREEN}{$\uparrow$} & 56.67 \textcolor{GREEN}{$\uparrow$} & 66.82 \textcolor{GREEN}{$\uparrow$} & 60.24 \textcolor{GREEN}{$\uparrow$} & 46.44 \textcolor{GREEN}{$\uparrow$} & 55.84 \textcolor{GREEN}{$\uparrow$} & \underline{49.54} \textcolor{GREEN}{$\uparrow$} \\
    \rowcolor{gray!10} Qwen3-14B & Non-Reasoning & 34.73 & 42.00 & 36.67 & 29.77 & 44.00 & 32.32 & 21.89 & 25.75 & 21.31 & 26.75 & 38.56 & 29.58 & 28.29 & 37.58 & 29.97 \\
     & FTRL-Reinforce++ & 36.92 \textcolor{GREEN}{$\uparrow$} & \underline{50.00} \textcolor{GREEN}{$\uparrow$} & 40.35 \textcolor{GREEN}{$\uparrow$} & 33.59 \textcolor{GREEN}{$\uparrow$} & \underline{51.33} \textcolor{GREEN}{$\uparrow$} & 37.89 \textcolor{GREEN}{$\uparrow$} & 48.62 \textcolor{GREEN}{$\uparrow$} & \textbf{62.47} \textcolor{GREEN}{$\uparrow$} & 52.57 \textcolor{GREEN}{$\uparrow$} & 57.12 \textcolor{GREEN}{$\uparrow$} & \underline{73.38} \textcolor{GREEN}{$\uparrow$} & 62.16 \textcolor{GREEN}{$\uparrow$} & 44.06 \textcolor{GREEN}{$\uparrow$} & \textbf{59.30} \textcolor{GREEN}{$\uparrow$} & 48.24 \textcolor{GREEN}{$\uparrow$} \\
     & FTRL-GRPO & 42.33 \textcolor{GREEN}{$\uparrow$} & \textbf{52.00} \textcolor{GREEN}{$\uparrow$} & \textbf{45.33} \textcolor{GREEN}{$\uparrow$} & 37.73 \textcolor{GREEN}{$\uparrow$} & 48.00 \textcolor{GREEN}{$\uparrow$} & \textbf{41.35} \textcolor{GREEN}{$\uparrow$} & 51.84 \textcolor{GREEN}{$\uparrow$} & 58.43 \textcolor{GREEN}{$\uparrow$} & \textbf{54.02} \textcolor{GREEN}{$\uparrow$} & 65.71 \textcolor{GREEN}{$\uparrow$} & 68.42 \textcolor{GREEN}{$\uparrow$} & \underline{66.56} \textcolor{GREEN}{$\uparrow$} & 49.41 \textcolor{GREEN}{$\uparrow$} & \underline{56.71} \textcolor{GREEN}{$\uparrow$} & \textbf{51.82} \textcolor{GREEN}{$\uparrow$} \\
    \rowcolor{gray!10} Qwen3-8B & Reasoning & 23.33 & 28.00 & 24.67 & 25.59 & 37.67 & 24.93 & 25.43 & 37.84 & 28.00 & 36.76 & 50.00 & 40.47 & 27.78 & 38.38 & 29.52 \\
     & FTRL-Reinforce++ & 28.00 \textcolor{GREEN}{$\uparrow$} & 36.00 \textcolor{GREEN}{$\uparrow$} & 30.33 \textcolor{GREEN}{$\uparrow$} & 31.60 \textcolor{GREEN}{$\uparrow$} & 36.67 \textcolor{GREEN}{$\uparrow$} & 31.99 \textcolor{GREEN}{$\uparrow$} & 36.84 \textcolor{GREEN}{$\uparrow$} & 43.16 \textcolor{GREEN}{$\uparrow$} & 38.69 \textcolor{GREEN}{$\uparrow$} & 56.02 \textcolor{GREEN}{$\uparrow$} & 58.96 \textcolor{GREEN}{$\uparrow$} & 56.61 \textcolor{GREEN}{$\uparrow$} & 38.11 \textcolor{GREEN}{$\uparrow$} & 43.70 \textcolor{GREEN}{$\uparrow$} & 39.40 \textcolor{GREEN}{$\uparrow$} \\
     & FTRL-GRPO & 25.67 \textcolor{GREEN}{$\uparrow$} & 30.00 \textcolor{GREEN}{$\uparrow$} & 27.00 \textcolor{GREEN}{$\uparrow$} & 37.83 \textcolor{GREEN}{$\uparrow$} & 44.67 \textcolor{GREEN}{$\uparrow$} & 39.07 \textcolor{GREEN}{$\uparrow$} & 43.07 \textcolor{GREEN}{$\uparrow$} & 37.41 \textcolor{RED}{$\downarrow$} & 37.49 \textcolor{GREEN}{$\uparrow$} & 56.11 \textcolor{GREEN}{$\uparrow$} & 48.18 \textcolor{GREEN}{$\uparrow$} & 50.75 \textcolor{GREEN}{$\uparrow$} & 40.67 \textcolor{GREEN}{$\uparrow$} & 40.06 \textcolor{GREEN}{$\uparrow$} & 38.58 \textcolor{GREEN}{$\uparrow$} \\
    \rowcolor{gray!10} Qwen3-14B & Reasoning & 25.17 & 32.00 & 27.13 & 25.32 & 40.33 & 29.05 & 33.08 & 37.74 & 30.79 & 44.07 & 50.86 & 44.85 & 31.91 & 40.23 & 32.96 \\
     & FTRL-Reinforce++ & 34.83 \textcolor{GREEN}{$\uparrow$} & 42.00 \textcolor{GREEN}{$\uparrow$} & 35.47 \textcolor{GREEN}{$\uparrow$} & 31.96 \textcolor{GREEN}{$\uparrow$} & 42.00 \textcolor{GREEN}{$\uparrow$} & 33.78 \textcolor{GREEN}{$\uparrow$} & 42.67 \textcolor{GREEN}{$\uparrow$} & 43.03 \textcolor{GREEN}{$\uparrow$} & 41.32 \textcolor{GREEN}{$\uparrow$} & \underline{67.15} \textcolor{GREEN}{$\uparrow$} & 65.40 \textcolor{GREEN}{$\uparrow$} & 65.59 \textcolor{GREEN}{$\uparrow$} & 44.15 \textcolor{GREEN}{$\uparrow$} & 48.11 \textcolor{GREEN}{$\uparrow$} & 44.04 \textcolor{GREEN}{$\uparrow$} \\
     & FTRL-GRPO & 31.67 \textcolor{GREEN}{$\uparrow$} & 40.00 \textcolor{GREEN}{$\uparrow$} & 34.33 \textcolor{GREEN}{$\uparrow$} & 30.18 \textcolor{GREEN}{$\uparrow$} & 43.00 \textcolor{GREEN}{$\uparrow$} & 34.57 \textcolor{GREEN}{$\uparrow$} & 44.64 \textcolor{GREEN}{$\uparrow$} & 49.55 \textcolor{GREEN}{$\uparrow$} & 46.00 \textcolor{GREEN}{$\uparrow$} & 60.19 \textcolor{GREEN}{$\uparrow$} & 58.12 \textcolor{GREEN}{$\uparrow$} & 58.52 \textcolor{GREEN}{$\uparrow$} & 41.67 \textcolor{GREEN}{$\uparrow$} & 47.67 \textcolor{GREEN}{$\uparrow$} & 43.36 \textcolor{GREEN}{$\uparrow$} \\
    \bottomrule
    \end{tabular}
    }
    \caption{Detailed evaluation results on our own dataset across different scenarios. The best result for each scenario is \textbf{bolded}, and the second-best is \underline{underlined}. Performance improvements over the base model after training with our method (i.e., FTRL-Reinforce++ and FTRL-GRPO) are indicated with \textcolor{GREEN}{$\uparrow$}, while performance declines are marked with \textcolor{RED}{$\downarrow$}.}
    \label{tab:self-result}
\end{table*}

\clearpage

\begin{table*}[h]
    \centering
    \resizebox{0.77\linewidth}{!}
    {
    \begin{tabular}{ll l l l l}
    \toprule
    \multirow{2}*{\textbf{Family}} & \multirow{2}*{\textbf{Version}} & \multicolumn{1}{l}{\textbf{Direct}} & \multicolumn{1}{l}{\textbf{Mandatory}} & \multicolumn{1}{l}{\textbf{Free}} & \multicolumn{1}{l}{\textbf{Avg}} \\ 
    \cmidrule(l){3-3} \cmidrule(l){4-4} \cmidrule(l){5-5} \cmidrule(l){6-6}
     & & \textbf{AC} & \textbf{AC} & \textbf{AC} & \textbf{AC} \\
    \midrule
    \multicolumn{6}{c}{\textit{Closed-Source LLMs}} \\
    Gemini & 2.5-Flash & 49.55 & 35.18 & 33.37 & 39.37 \\
    Gemini & 2.5-Pro & \underline{54.07} & 38.99 & 42.91 & 45.32 \\
    Claude & 4.0-Sonnet & \textbf{55.48} & 49.35 & \textbf{57.09} & \textbf{53.97} \\
    GPT & 3.5-Turbo & 13.37 & 31.56 & 32.26 & 25.73 \\
    GPT & 4o & 18.99 & 44.32 & 45.63 & 36.31 \\
    \midrule
    \multicolumn{6}{c}{\textit{Open-Source LLMs (32B - 72B)}} \\
    Qwen2.5-32B & Instruct & 14.57 & 24.22 & 20.30 & 19.70 \\
    Qwen2.5-72B & Instruct & 12.76 & 43.02 & 35.08 & 30.29 \\
    Qwen3-32B & Non-Reasoning & 13.77 & 47.54 & 44.72 & 35.34 \\
    Qwen3-32B & Reasoning & 38.89 & \textbf{55.58} & \underline{55.88} & \underline{50.12} \\
    \midrule
    \multicolumn{6}{c}{\textit{Open-Source LLMs (7B - 14B)}} \\
    \rowcolor{gray!10} Qwen2.5-7B & Instruct & 8.94 & 11.36 & 15.68 & 11.99 \\
     & FTRL-Reinforce++ & 8.74 \textcolor{RED}{$\downarrow$} & 30.95 \textcolor{GREEN}{$\uparrow$} & 30.65 \textcolor{GREEN}{$\uparrow$} & 23.45 \textcolor{GREEN}{$\uparrow$} \\
     & FTRL-GRPO & 6.93 \textcolor{RED}{$\downarrow$} & 44.32 \textcolor{GREEN}{$\uparrow$} & 37.79 \textcolor{GREEN}{$\uparrow$} & 29.68 \textcolor{GREEN}{$\uparrow$} \\
    \rowcolor{gray!10} Qwen2.5-14B & Instruct & 14.07 & 35.48 & 27.84 & 25.80 \\
     & FTRL-Reinforce++ & 13.37 \textcolor{RED}{$\downarrow$} & 47.94 \textcolor{GREEN}{$\uparrow$} & 47.94 \textcolor{GREEN}{$\uparrow$} & 36.42 \textcolor{GREEN}{$\uparrow$} \\
     & FTRL-GRPO & 14.37 \textcolor{GREEN}{$\uparrow$} & 37.49 \textcolor{GREEN}{$\uparrow$} & 28.04 \textcolor{GREEN}{$\uparrow$} & 26.63 \textcolor{GREEN}{$\uparrow$} \\
    \rowcolor{gray!10} Qwen3-8B & Non-Reasoning & 20.30 & 34.07 & 31.26 & 28.54 \\
     & FTRL-Reinforce++ & 23.02 \textcolor{GREEN}{$\uparrow$} & 46.23 \textcolor{GREEN}{$\uparrow$} & 44.12 \textcolor{GREEN}{$\uparrow$} & 37.79 \textcolor{GREEN}{$\uparrow$} \\
     & FTRL-GRPO & 20.80 \textcolor{GREEN}{$\uparrow$} & 48.74 \textcolor{GREEN}{$\uparrow$} & 47.24 \textcolor{GREEN}{$\uparrow$} & 38.93 \textcolor{GREEN}{$\uparrow$} \\
    \rowcolor{gray!10} Qwen3-14B & Non-Reasoning & 12.56 & 29.55 & 30.45 & 24.19 \\
     & FTRL-Reinforce++ & 14.57 \textcolor{GREEN}{$\uparrow$} & 50.15 \textcolor{GREEN}{$\uparrow$} & 50.25 \textcolor{GREEN}{$\uparrow$} & 38.32 \textcolor{GREEN}{$\uparrow$} \\
     & FTRL-GRPO & 13.07 \textcolor{GREEN}{$\uparrow$} & 50.25 \textcolor{GREEN}{$\uparrow$} & 51.46 \textcolor{GREEN}{$\uparrow$} & 38.26 \textcolor{GREEN}{$\uparrow$} \\
    \rowcolor{gray!10} Qwen3-8B & Reasoning & 32.66 & 46.43 & 43.02 & 40.70 \\
     & FTRL-Reinforce++ & 34.17 \textcolor{GREEN}{$\uparrow$} & 48.94 \textcolor{GREEN}{$\uparrow$} & 40.60 \textcolor{RED}{$\downarrow$} & 41.24 \textcolor{GREEN}{$\uparrow$} \\
     & FTRL-GRPO & 31.76 \textcolor{RED}{$\downarrow$} & \underline{51.06} \textcolor{GREEN}{$\uparrow$} & 38.69 \textcolor{RED}{$\downarrow$} & 40.50 \textcolor{RED}{$\downarrow$} \\
    \rowcolor{gray!10} Qwen3-14B & Reasoning & 36.68 & 42.21 & 41.41 & 40.10 \\
     & FTRL-Reinforce++ & 38.89 \textcolor{GREEN}{$\uparrow$} & 50.75 \textcolor{GREEN}{$\uparrow$} & 44.72 \textcolor{GREEN}{$\uparrow$} & 44.79 \textcolor{GREEN}{$\uparrow$} \\
     & FTRL-GRPO & 35.48 \textcolor{RED}{$\downarrow$} & 47.34 \textcolor{GREEN}{$\uparrow$} & 42.11 \textcolor{GREEN}{$\uparrow$} & 41.64 \textcolor{GREEN}{$\uparrow$} \\
    \bottomrule
    \end{tabular}
    }
    \caption{Detailed evaluation results on ToolHop across different scenarios. The best result for each scenario is \textbf{bolded}, and the second-best is \underline{underlined}. Performance improvements over the base model after training with our method (i.e., FTRL-Reinforce++ and FTRL-GRPO) are indicated with \textcolor{GREEN}{$\uparrow$}, while performance declines are marked with \textcolor{RED}{$\downarrow$}.}
    \label{tab:toolhop_result}
\end{table*}

\clearpage

\begin{table*}[h]
    \centering
    \resizebox{0.65\linewidth}{!}
    {
    \begin{tabular}{ll l l l}
    \toprule
    \multirow{2}*{\textbf{Family}} & \multirow{2}*{\textbf{Version}} & \textbf{Retail} & \textbf{Airline} & \textbf{Avg} \\
    \cmidrule(l){3-3} \cmidrule(l){4-4} \cmidrule(l){5-5}
     &  & \textbf{Pass\texttt{\char`^}1} & \textbf{Pass\texttt{\char`^}1} & \textbf{Pass\texttt{\char`^}1} \\
    \midrule
    \multicolumn{5}{c}{\textit{Closed-Source LLMs}} \\
    Gemini & 2.5-Flash & 46.09 & 34.00 & 40.04 \\
    Gemini & 2.5-Pro & \textbf{52.17} & \underline{42.00} & \underline{47.09} \\
    Claude & 4.0-Sonnet & \underline{50.43} & \textbf{50.00} & \textbf{50.22} \\
    GPT & 3.5-Turbo & 18.26 & 12.00 & 15.13 \\
    GPT & 4o & 40.87 & 34.00 & 37.43 \\
    \midrule
    \multicolumn{5}{c}{\textit{Open-Source LLMs (32B - 72B)}} \\
    Qwen2.5-32B & Instruct & 27.83 & 16.00 & 21.91 \\
    Qwen2.5-72B & Instruct & 36.52 & 32.00 & 34.26 \\
    Qwen3-32B & Non-Reasoning & 34.78 & 20.00 & 27.39 \\
    Qwen3-32B & Reasoning & 40.00 & 22.00 & 31.00 \\
    \midrule
    \multicolumn{5}{c}{\textit{Open-Source LLMs (7B - 14B)}} \\
    \rowcolor{gray!10} Qwen2.5-7B & Instruct & 7.83 & 4.00 & 5.91 \\
     & FTRL-Reinforce++ & 7.83 & 16.00 \textcolor{GREEN}{$\uparrow$} & 11.91 \textcolor{GREEN}{$\uparrow$} \\
     & FTRL-GRPO & 7.83 & 6.00 \textcolor{GREEN}{$\uparrow$} & 6.91 \textcolor{GREEN}{$\uparrow$} \\
    \rowcolor{gray!10} Qwen2.5-14B & Instruct & 23.48 & 10.00 & 16.74 \\
     & FTRL-Reinforce++ & 35.65 \textcolor{GREEN}{$\uparrow$} & 18.00 \textcolor{GREEN}{$\uparrow$} & 26.83 \textcolor{GREEN}{$\uparrow$} \\
     & FTRL-GRPO & 40.87 \textcolor{GREEN}{$\uparrow$} & 10.00 & 25.43 \textcolor{GREEN}{$\uparrow$} \\
    \rowcolor{gray!10} Qwen3-8B & Non-Reasoning & 18.26 & 2.00 & 10.13 \\
     & FTRL-Reinforce++ & 33.91 \textcolor{GREEN}{$\uparrow$} & 10.00 \textcolor{GREEN}{$\uparrow$} & 21.96 \textcolor{GREEN}{$\uparrow$} \\
     & FTRL-GRPO & 28.70 \textcolor{GREEN}{$\uparrow$} & 18.00 \textcolor{GREEN}{$\uparrow$} & 23.35 \textcolor{GREEN}{$\uparrow$} \\
    \rowcolor{gray!10} Qwen3-14B & Non-Reasoning & 23.48 & 4.00 & 13.74 \\
     & FTRL-Reinforce++ & 25.22 \textcolor{GREEN}{$\uparrow$} & 10.00 \textcolor{GREEN}{$\uparrow$} & 17.61 \textcolor{GREEN}{$\uparrow$} \\
     & FTRL-GRPO & 36.52 \textcolor{GREEN}{$\uparrow$} & 12.00 \textcolor{GREEN}{$\uparrow$} & 24.26 \textcolor{GREEN}{$\uparrow$} \\
    \rowcolor{gray!10} Qwen3-8B & Reasoning & 20.87 & 12.00 & 16.43 \\
     & FTRL-Reinforce++ & 33.04 \textcolor{GREEN}{$\uparrow$} & 32.00 \textcolor{GREEN}{$\uparrow$} & 32.52 \textcolor{GREEN}{$\uparrow$} \\
     & FTRL-GRPO & 27.83 \textcolor{GREEN}{$\uparrow$} & 30.00 \textcolor{GREEN}{$\uparrow$} & 28.91 \textcolor{GREEN}{$\uparrow$} \\
    \rowcolor{gray!10} Qwen3-14B & Reasoning & 21.74 & 16.00 & 18.87 \\
     & FTRL-Reinforce++ & 32.17 \textcolor{GREEN}{$\uparrow$} & 22.00 \textcolor{GREEN}{$\uparrow$} & 27.09 \textcolor{GREEN}{$\uparrow$} \\
     & FTRL-GRPO & 37.39 \textcolor{GREEN}{$\uparrow$} & 26.00 \textcolor{GREEN}{$\uparrow$} & 31.70 \textcolor{GREEN}{$\uparrow$} \\
    \bottomrule
    \end{tabular}
    }
    \caption{Detailed evaluation results on $\tau$-bench across different scenarios. The best result for each scenario is \textbf{bolded}, and the second-best is \underline{underlined}. Performance improvements over the base model after training with our method (i.e., FTRL-Reinforce++ and FTRL-GRPO) are indicated with \textcolor{GREEN}{$\uparrow$}, while performance declines are marked with \textcolor{RED}{$\downarrow$}.}
    \label{tab:tau_bench_result}
\end{table*}

\clearpage

\begin{table*}[h]
    \centering
    \resizebox{\textwidth}{!}{
    \begin{tabular}{ll lll lll lll lll lll lll}
    \toprule
    \multirow{2}*{\textbf{Family}} & \multirow{2}*{\textbf{Version}} & \multicolumn{3}{l}{\textbf{Clean}} & \multicolumn{3}{l}{\textbf{Slight}} & \multicolumn{3}{l}{\textbf{Medium}} & \multicolumn{3}{l}{\textbf{Heavy}} & \multicolumn{3}{l}{\textbf{Union}} & \multicolumn{3}{l}{\textbf{Avg}} \\ 
    \cmidrule(l){3-5} \cmidrule(l){6-8} \cmidrule(l){9-11} \cmidrule(l){12-14} \cmidrule(l){15-17} \cmidrule(l){18-20}
     & & \textbf{TS} & \textbf{PI} & \textbf{CF} & \textbf{TS} & \textbf{PI} & \textbf{CF} & \textbf{TS} & \textbf{PI} & \textbf{CF} & \textbf{TS} & \textbf{PI} & \textbf{CF} & \textbf{TS} & \textbf{PI} & \textbf{CF} & \textbf{TS} & \textbf{PI} & \textbf{CF} \\    
    \midrule
    \multicolumn{20}{c}{\textit{Closed-Source LLMs}} \\
    Gemini & 2.5-Flash & 58.10 & 37.14 & 25.71 & 56.67 & 40.00 & 26.19 & 62.86 & 39.52 & 25.24 & 42.86 & 26.67 & 17.62 & 47.62 & 31.43 & 19.05 & 53.62 & 34.95 & 22.76 \\
    Gemini & 2.5-Pro & 49.52 & 45.71 & 30.48 & 50.00 & 44.29 & 25.24 & 58.10 & \underline{52.86} & \underline{34.76} & 38.10 & 28.10 & 18.57 & 49.52 & 40.95 & \underline{28.57} & 49.05 & 42.38 & 27.52 \\
    Claude & 4.0-Sonnet & 34.29 & 31.43 & 18.10 & 45.24 & 40.00 & 24.76 & 36.19 & 32.86 & 21.43 & 38.10 & 29.52 & 17.14 & 44.76 & 40.95 & 27.62 & 39.72 & 34.95 & 21.81 \\
    GPT & 3.5-Turbo & 75.24 & 33.33 & 19.05 & 65.71 & 27.14 & 16.67 & 73.33 & 36.67 & 21.43 & 58.10 & 23.33 & 13.33 & 56.19 & 28.57 & 16.19 & 65.71 & 29.81 & 17.33 \\
    GPT & 4o & \textbf{85.71} & \underline{50.48} & \underline{31.43} & \textbf{84.29} & \textbf{51.90} & \textbf{33.81} & \textbf{84.76} & \textbf{57.14} & \textbf{36.19} & 64.76 & \textbf{40.00} & \textbf{25.71} & \underline{74.29} & \textbf{47.62} & \textbf{30.48} & \textbf{78.76} & \textbf{49.43} & \textbf{31.52} \\
    \midrule
    \multicolumn{20}{c}{\textit{Open-Source LLMs (32B - 72B)}} \\
    Qwen2.5-32B & Instruct & 81.90 & 42.86 & 22.86 & 81.43 & 42.38 & 22.38 & 80.48 & 38.10 & 22.38 & 60.95 & 30.48 & 18.10 & 72.38 & 30.48 & 14.29 & 75.43 & 36.86 & 20.00 \\
    Qwen2.5-72B & Instruct & 80.00 & 26.67 & 18.10 & 79.05 & 29.05 & 19.05 & 76.67 & 28.10 & 18.57 & 55.71 & 21.90 & 13.81 & 69.52 & 25.71 & 16.19 & 72.19 & 26.29 & 17.14 \\
    Qwen3-32B & Non-Reasoning & 77.14 & 40.00 & 22.86 & 78.10 & 45.71 & 26.67 & 78.57 & 49.52 & 30.48 & 60.95 & 34.29 & 21.43 & 71.43 & 39.05 & 22.86 & 73.24 & 41.71 & 24.86 \\
    Qwen3-32B & Reasoning & 57.14 & 36.19 & 18.10 & 60.95 & 38.10 & 24.29 & 60.00 & 37.62 & 24.76 & 44.76 & 29.52 & 16.67 & 46.67 & 27.62 & 17.14 & 53.90 & 33.81 & 20.19 \\
    \midrule
    \multicolumn{20}{c}{\textit{Open-Source LLMs (7B - 14B)}} \\
    \rowcolor{gray!10} Qwen2.5-7B & Instruct & 78.10 & 31.43 & 16.19 & 73.81 & 26.19 & 14.76 & 73.33 & 29.52 & 16.19 & 59.05 & 23.81 & 13.81 & 67.62 & 25.71 & 16.19 & 70.38 & 27.33 & 15.43 \\
     & FTRL-Reinforce++ & 79.05 \textcolor{GREEN}{$\uparrow$} & 48.57 \textcolor{GREEN}{$\uparrow$} & 30.48 \textcolor{GREEN}{$\uparrow$} & 72.86 \textcolor{RED}{$\downarrow$} & 42.38 \textcolor{GREEN}{$\uparrow$} & 28.10 \textcolor{GREEN}{$\uparrow$} & 73.81 \textcolor{GREEN}{$\uparrow$} & 41.90 \textcolor{GREEN}{$\uparrow$} & 25.71 \textcolor{GREEN}{$\uparrow$} & 63.33 \textcolor{GREEN}{$\uparrow$} & 37.62 \textcolor{GREEN}{$\uparrow$} & 23.33 \textcolor{GREEN}{$\uparrow$} & 65.71 \textcolor{RED}{$\downarrow$} & 34.29 \textcolor{GREEN}{$\uparrow$} & 22.86 \textcolor{GREEN}{$\uparrow$} & 70.95 \textcolor{GREEN}{$\uparrow$} & 40.95 \textcolor{GREEN}{$\uparrow$} & 26.10 \textcolor{GREEN}{$\uparrow$} \\
     & FTRL-GRPO & 80.95 \textcolor{GREEN}{$\uparrow$} & 34.29 \textcolor{GREEN}{$\uparrow$} & 20.95 \textcolor{GREEN}{$\uparrow$} & 77.62 \textcolor{GREEN}{$\uparrow$} & 31.90 \textcolor{GREEN}{$\uparrow$} & 20.48 \textcolor{GREEN}{$\uparrow$} & 75.24 \textcolor{GREEN}{$\uparrow$} & 32.38 \textcolor{GREEN}{$\uparrow$} & 19.52 \textcolor{GREEN}{$\uparrow$} & 64.76 \textcolor{GREEN}{$\uparrow$} & 30.00 \textcolor{GREEN}{$\uparrow$} & 19.52 \textcolor{GREEN}{$\uparrow$} & 71.43 \textcolor{GREEN}{$\uparrow$} & 25.71 & 14.29 \textcolor{RED}{$\downarrow$} & 74.00 \textcolor{GREEN}{$\uparrow$} & 30.86 \textcolor{GREEN}{$\uparrow$} & 18.95 \textcolor{GREEN}{$\uparrow$} \\
    \rowcolor{gray!10} Qwen2.5-14B & Instruct & \underline{83.81} & 47.62 & 28.57 & 77.14 & 40.95 & 23.33 & 76.19 & 42.86 & 25.24 & 57.14 & 30.48 & 17.14 & 64.76 & 29.52 & 12.38 & 71.81 & 38.29 & 21.33 \\
     & FTRL-Reinforce++ & \underline{83.81} & \textbf{55.24} \textcolor{GREEN}{$\uparrow$} & \textbf{35.24} \textcolor{GREEN}{$\uparrow$} & 80.00 \textcolor{GREEN}{$\uparrow$} & \underline{49.05} \textcolor{GREEN}{$\uparrow$} & 30.48 \textcolor{GREEN}{$\uparrow$} & 76.67 \textcolor{GREEN}{$\uparrow$} & 49.05 \textcolor{GREEN}{$\uparrow$} & 31.90 \textcolor{GREEN}{$\uparrow$} & 60.00 \textcolor{GREEN}{$\uparrow$} & 33.81 \textcolor{GREEN}{$\uparrow$} & 21.43 \textcolor{GREEN}{$\uparrow$} & 64.76 & 39.05 \textcolor{GREEN}{$\uparrow$} & 21.90 \textcolor{GREEN}{$\uparrow$} & 73.05 \textcolor{GREEN}{$\uparrow$} & \underline{45.24} \textcolor{GREEN}{$\uparrow$} & \underline{28.19} \textcolor{GREEN}{$\uparrow$} \\
     & FTRL-GRPO & 76.19 \textcolor{RED}{$\downarrow$} & 39.05 \textcolor{RED}{$\downarrow$} & 23.81 \textcolor{RED}{$\downarrow$} & 76.19 \textcolor{RED}{$\downarrow$} & 38.57 \textcolor{RED}{$\downarrow$} & 24.29 \textcolor{GREEN}{$\uparrow$} & 77.62 \textcolor{GREEN}{$\uparrow$} & 39.05 \textcolor{RED}{$\downarrow$} & 23.33 \textcolor{RED}{$\downarrow$} & 60.00 \textcolor{GREEN}{$\uparrow$} & 30.95 \textcolor{GREEN}{$\uparrow$} & 17.62 \textcolor{GREEN}{$\uparrow$} & 67.62 \textcolor{GREEN}{$\uparrow$} & 33.33 \textcolor{GREEN}{$\uparrow$} & 19.05 \textcolor{GREEN}{$\uparrow$} & 71.52 \textcolor{RED}{$\downarrow$} & 36.19 \textcolor{RED}{$\downarrow$} & 21.62 \textcolor{GREEN}{$\uparrow$} \\
    \rowcolor{gray!10} Qwen3-8B & Non-Reasoning & 79.05 & 35.24 & 20.00 & 80.95 & 40.00 & 26.67 & 80.48 & 37.14 & 22.38 & 65.71 & 32.86 & 20.00 & 71.43 & 36.19 & 21.90 & 75.52 & 36.29 & 22.19 \\
     & FTRL-Reinforce++ & \underline{83.81} \textcolor{GREEN}{$\uparrow$} & 42.86 \textcolor{GREEN}{$\uparrow$} & 25.71 \textcolor{GREEN}{$\uparrow$} & 81.43 \textcolor{GREEN}{$\uparrow$} & 46.67 \textcolor{GREEN}{$\uparrow$} & 29.05 \textcolor{GREEN}{$\uparrow$} & 81.90 \textcolor{GREEN}{$\uparrow$} & 44.76 \textcolor{GREEN}{$\uparrow$} & 25.71 \textcolor{GREEN}{$\uparrow$} & \textbf{68.10} \textcolor{GREEN}{$\uparrow$} & 36.67 \textcolor{GREEN}{$\uparrow$} & 23.33 \textcolor{GREEN}{$\uparrow$} & \textbf{75.24} \textcolor{GREEN}{$\uparrow$} & 41.90 \textcolor{GREEN}{$\uparrow$} & 25.71 \textcolor{GREEN}{$\uparrow$} & \underline{78.10} \textcolor{GREEN}{$\uparrow$} & 42.57 \textcolor{GREEN}{$\uparrow$} & 25.90 \textcolor{GREEN}{$\uparrow$} \\
     & FTRL-GRPO & 82.86 \textcolor{GREEN}{$\uparrow$} & 47.62 \textcolor{GREEN}{$\uparrow$} & 27.62 \textcolor{GREEN}{$\uparrow$} & \underline{83.33} \textcolor{GREEN}{$\uparrow$} & 48.10 \textcolor{GREEN}{$\uparrow$} & \underline{32.38} \textcolor{GREEN}{$\uparrow$} & 82.38 \textcolor{GREEN}{$\uparrow$} & 44.76 \textcolor{GREEN}{$\uparrow$} & 25.71 \textcolor{GREEN}{$\uparrow$} & 66.19 & \underline{38.10} \textcolor{GREEN}{$\uparrow$} & 21.90 \textcolor{GREEN}{$\uparrow$} & 73.33 \textcolor{GREEN}{$\uparrow$} & \underline{44.76} \textcolor{GREEN}{$\uparrow$} & 27.62 \textcolor{GREEN}{$\uparrow$} & 77.62 \textcolor{GREEN}{$\uparrow$} & 44.67 \textcolor{GREEN}{$\uparrow$} & 27.05 \textcolor{GREEN}{$\uparrow$} \\
    \rowcolor{gray!10} Qwen3-14B & Non-Reasoning & 81.90 & 38.10 & 22.86 & 82.86 & 39.52 & 20.95 & 82.86 & 40.95 & 21.43 & 62.38 & 30.95 & 20.00 & 69.52 & 32.38 & 18.10 & 75.90 & 36.38 & 20.67 \\
     & FTRL-Reinforce++ & 82.86 \textcolor{GREEN}{$\uparrow$} & 45.71 \textcolor{GREEN}{$\uparrow$} & 26.67 \textcolor{GREEN}{$\uparrow$} & 81.43 \textcolor{RED}{$\downarrow$} & 44.29 \textcolor{GREEN}{$\uparrow$} & 26.19 \textcolor{GREEN}{$\uparrow$} & \underline{83.81} \textcolor{GREEN}{$\uparrow$} & 47.14 \textcolor{GREEN}{$\uparrow$} & 27.14 \textcolor{GREEN}{$\uparrow$} & \underline{67.62} \textcolor{GREEN}{$\uparrow$} & 37.62 \textcolor{GREEN}{$\uparrow$} & \underline{24.29} \textcolor{GREEN}{$\uparrow$} & 73.33 \textcolor{GREEN}{$\uparrow$} & 38.10 \textcolor{GREEN}{$\uparrow$} & 21.90 \textcolor{GREEN}{$\uparrow$} & 77.81 \textcolor{GREEN}{$\uparrow$} & 42.57 \textcolor{GREEN}{$\uparrow$} & 25.24 \textcolor{GREEN}{$\uparrow$} \\
     & FTRL-GRPO & 78.10 \textcolor{RED}{$\downarrow$} & 43.81 \textcolor{GREEN}{$\uparrow$} & 22.86 & 80.00 \textcolor{RED}{$\downarrow$} & 43.81 \textcolor{GREEN}{$\uparrow$} & 24.29 \textcolor{GREEN}{$\uparrow$} & 80.00 \textcolor{RED}{$\downarrow$} & 43.81 \textcolor{GREEN}{$\uparrow$} & 23.33 \textcolor{GREEN}{$\uparrow$} & 64.29 \textcolor{GREEN}{$\uparrow$} & 34.29 \textcolor{GREEN}{$\uparrow$} & 20.00 & \underline{74.29} \textcolor{GREEN}{$\uparrow$} & 39.05 \textcolor{GREEN}{$\uparrow$} & 21.90 \textcolor{GREEN}{$\uparrow$} & 75.33 \textcolor{RED}{$\downarrow$} & 40.95 \textcolor{GREEN}{$\uparrow$} & 22.48 \textcolor{GREEN}{$\uparrow$} \\
    \rowcolor{gray!10} Qwen3-8B & Reasoning & 62.86 & 36.19 & 23.81 & 51.43 & 32.86 & 20.00 & 55.71 & 38.57 & 21.43 & 46.67 & 29.52 & 14.76 & 53.33 & 34.29 & 21.90 & 54.00 & 34.29 & 20.38 \\
     & FTRL-Reinforce++ & 59.05 \textcolor{RED}{$\downarrow$} & 40.95 \textcolor{GREEN}{$\uparrow$} & 25.71 \textcolor{GREEN}{$\uparrow$} & 57.14 \textcolor{GREEN}{$\uparrow$} & 43.81 \textcolor{GREEN}{$\uparrow$} & 25.71 \textcolor{GREEN}{$\uparrow$} & 56.67 \textcolor{GREEN}{$\uparrow$} & 41.43 \textcolor{GREEN}{$\uparrow$} & 24.76 \textcolor{GREEN}{$\uparrow$} & 49.05 \textcolor{GREEN}{$\uparrow$} & 30.00 \textcolor{GREEN}{$\uparrow$} & 18.10 \textcolor{GREEN}{$\uparrow$} & 45.71 \textcolor{RED}{$\downarrow$} & 31.43 \textcolor{RED}{$\downarrow$} & 17.14 \textcolor{RED}{$\downarrow$} & 53.52 \textcolor{RED}{$\downarrow$} & 37.52 \textcolor{GREEN}{$\uparrow$} & 22.29 \textcolor{GREEN}{$\uparrow$} \\
     & FTRL-GRPO & 60.00 \textcolor{RED}{$\downarrow$} & 41.90 \textcolor{GREEN}{$\uparrow$} & 23.81 & 57.14 \textcolor{GREEN}{$\uparrow$} & 42.86 \textcolor{GREEN}{$\uparrow$} & 26.19 \textcolor{GREEN}{$\uparrow$} & 54.76 \textcolor{RED}{$\downarrow$} & 43.33 \textcolor{GREEN}{$\uparrow$} & 25.24 \textcolor{GREEN}{$\uparrow$} & 49.52 \textcolor{GREEN}{$\uparrow$} & 30.00 \textcolor{GREEN}{$\uparrow$} & 16.67 \textcolor{GREEN}{$\uparrow$} & 51.43 \textcolor{RED}{$\downarrow$} & 38.10 \textcolor{GREEN}{$\uparrow$} & 22.86 \textcolor{GREEN}{$\uparrow$} & 54.57 \textcolor{GREEN}{$\uparrow$} & 39.24 \textcolor{GREEN}{$\uparrow$} & 22.95 \textcolor{GREEN}{$\uparrow$} \\
    \rowcolor{gray!10} Qwen3-14B & Reasoning & 64.76 & 40.95 & 20.00 & 61.90 & 40.00 & 25.24 & 60.48 & 38.10 & 20.48 & 50.95 & 31.43 & 20.48 & 48.57 & 30.48 & 15.24 & 57.33 & 36.19 & 20.29 \\
     & FTRL-Reinforce++ & 67.62 \textcolor{GREEN}{$\uparrow$} & 43.81 \textcolor{GREEN}{$\uparrow$} & 26.67 \textcolor{GREEN}{$\uparrow$} & 57.62 \textcolor{RED}{$\downarrow$} & 38.57 \textcolor{RED}{$\downarrow$} & 20.95 \textcolor{RED}{$\downarrow$} & 61.90 \textcolor{GREEN}{$\uparrow$} & 40.00 \textcolor{GREEN}{$\uparrow$} & 20.95 \textcolor{GREEN}{$\uparrow$} & 52.38 \textcolor{GREEN}{$\uparrow$} & 30.48 \textcolor{RED}{$\downarrow$} & 19.52 \textcolor{RED}{$\downarrow$} & 55.24 \textcolor{GREEN}{$\uparrow$} & 33.33 \textcolor{GREEN}{$\uparrow$} & 20.00 \textcolor{GREEN}{$\uparrow$} & 58.95 \textcolor{GREEN}{$\uparrow$} & 37.24 \textcolor{GREEN}{$\uparrow$} & 21.62 \textcolor{GREEN}{$\uparrow$} \\
     & FTRL-GRPO & 67.62 \textcolor{GREEN}{$\uparrow$} & 40.95 & 20.95 \textcolor{GREEN}{$\uparrow$} & 62.86 \textcolor{GREEN}{$\uparrow$} & 41.90 \textcolor{GREEN}{$\uparrow$} & 23.81 \textcolor{RED}{$\downarrow$} & 67.14 \textcolor{GREEN}{$\uparrow$} & 43.81 \textcolor{GREEN}{$\uparrow$} & 26.19 \textcolor{GREEN}{$\uparrow$} & 51.90 \textcolor{GREEN}{$\uparrow$} & 31.43 & 18.10 \textcolor{RED}{$\downarrow$} & 53.33 \textcolor{GREEN}{$\uparrow$} & 32.38 \textcolor{GREEN}{$\uparrow$} & 18.10 \textcolor{GREEN}{$\uparrow$} & 60.57 \textcolor{GREEN}{$\uparrow$} & 38.10 \textcolor{GREEN}{$\uparrow$} & 21.43 \textcolor{GREEN}{$\uparrow$} \\
    \bottomrule
    \end{tabular}
    }
    \caption{Detailed evaluation results on RoTBench across different scenarios. The best result for each scenario is \textbf{bolded}, and the second-best is \underline{underlined}. Performance improvements over the base model after training with our method (i.e., FTRL-Reinforce++ and FTRL-GRPO) are indicated with \textcolor{GREEN}{$\uparrow$}, while performance declines are marked with \textcolor{RED}{$\downarrow$}.}
    \label{tab:rotbench-result}
\end{table*}

\clearpage

\begin{table*}[h]
\centering
\begin{tabular}{p{0.45\linewidth}|p{0.45\linewidth}}
\toprule
\rowcolor{gray!10}  \multicolumn{1}{c|}{\textbf{Qwen2.5-7B-Instruct}} & \multicolumn{1}{c}{\textbf{Qwen2.5-7B-FTRL-GRPO}} \\
\midrule
\multicolumn{2}{p{0.9\linewidth}}{\textbf{\textit{Question:}} Who succeeded Emir Nawaf Al-Ahmad Al-Jaber Al-Sabah after his death on December 16, 2023?} \\
\multicolumn{2}{p{0.9\linewidth}}{\textbf{\textit{Answer:}} Mishal Al-Ahmad Al-Jaber Al-Sabah} \\
\midrule
 \textbf{\textit{Called Tool:}} political\_successor\_finder & \textbf{\textit{Called Tool:}} political\_successor\_finder  \\
\midrule
\textbf{\textit{Filled Parameters:}} \{``political\_figure'': \textcolor{red}{``Nawaf Al-Ahmad Al-Jaber Al-Sabah''}, ``role'': ``Emir'', ``date\_of\_transition'': ``2023-12-16'', ``country'': ``Kuwait'', ``political\_system'': ``monarchy'', ``government\_level'': ``national''\} & \textbf{\textit{Filled Parameters:}} \{``political\_figure'': ``Emir Nawaf Al-Ahmad Al-Jaber Al-Sabah'', ``role'': ``Emir'', ``date\_of\_transition'': ``2023-12-16'', ``country'': ``Kuwait'', ``political\_system'': ``monarchy''\} \\
\midrule
\textbf{\textit{Environment Feedback:}} \textcolor{red}{No successor found for the given political figure and date.} & \textbf{\textit{Environment Feedback:}} \textcolor{green!60!black}{Mishal Al-Ahmad Al-Jaber Al-Sabah.}\\
\midrule
\textbf{\textit{Final Output:}} It seems there is \textcolor{red}{no recorded successor} for Emir Nawaf Al-Ahmad Al-Jaber Al-Sabah as of December 16, 2023. This might be due to the fact that the transition has not been officially announced or documented yet. Let's try again with a slightly later date to see if any information becomes available. & \textbf{\textit{Final Output:}} The successor to Emir Nawaf Al-Ahmad Al-Jaber Al-Sabah after his death on December 16, 2023, is \textcolor{green!60!black}{Mishal Al-Ahmad Al-Jaber Al-Sabah}. \\
\midrule
\multicolumn{2}{p{0.9\linewidth}}{\textbf{\textit{Analysis:}} Qwen2.5-7B-Instruct failed to accurately extract information from the user input due to contextual oversights, leading to incomplete parameter filling and, ultimately, an incorrect response. In contrast, the FTRL-GRPO variant maintained full contextual fidelity and successfully answered the question.} \\
\bottomrule
\end{tabular}
\caption{Performance comparison between Qwen2.5-7B-Instruct and Qwen2.5-7B-FTRL-GRPO.}
\label{tab:qwen2.5-7b-ftrl-grpo-vs-qwen2.5-7b}
\end{table*}

\clearpage

\begin{table*}[h]
\centering
\begin{tabular}{p{0.45\linewidth}|p{0.45\linewidth}}
\toprule
\rowcolor{gray!10} \multicolumn{1}{c|}{\textbf{Qwen3-8B-Reasoning}} & \multicolumn{1}{c}{\textbf{Qwen3-8B-Reasoning-FTRL-Reinforce++}} \\
\midrule
\multicolumn{2}{p{0.9\linewidth}}{\textbf{\textit{Question:}} How many seats did the SNS coalition win in the December 17, 2023 Serbian parliamentary election?} \\
\multicolumn{2}{p{0.9\linewidth}}{\textbf{\textit{Answer:}} 128} \\
\midrule
\textbf{\textit{Called Tool:}} election\_results\_analyzer & \textbf{\textit{Called Tool:}} election\_results\_analyzer \\
\midrule
\textbf{\textit{Filled Parameters:}} \{``election\_date'': ``2023-12-17'', ``country'': ``Serbia'', ``political\_entity'': \textcolor{red}{``SNS coalition''}, ``election\_type'': ``parliamentary'', ``data\_source'': ``official'', ``include\_contested'': false\} & \textbf{\textit{Filled Parameters:}} \{``election\_date'': ``2023-12-17'', ``country'': ``Serbia'', ``political\_entity'': ``SNS'', ``election\_type'': ``parliamentary'', ``data\_source'': ``media''\} \\
\midrule
\textbf{\textit{Environment Feedback:}} \textcolor{red}{No data available for the specified parameters.} & \textbf{\textit{Environment Feedback:}} \textcolor{green!60!black}{128} \\
\midrule
\textbf{\textit{Final Output:}} The provided tools do not have access to data for the specified election date (December 17, 2023) or the SNS coalition's results in Serbia. This could indicate an error in the date, as Serbia's last parliamentary election was held on \textcolor{red}{June 21, 2022}. For accurate results, please verify the election date or consult official Serbian electoral commission sources. & \textbf{\textit{Final Output:}} The SNS coalition won \textcolor{green!60!black}{128} seats in the December 17, 2023 Serbian parliamentary election. \\
\midrule
\multicolumn{2}{p{0.9\linewidth}}{\textbf{\textit{Analysis:}} Qwen3-8B-Reasoning struggles to correctly interpret user input and distinguish between entities, leading to incorrect parameter entries and failure to retrieve accurate information. In contrast, the FTRL-Reinforce++ version enhances the model’s contextual understanding and decision-making ability, allowing it to input the correct parameters and resolve the question effectively.} \\
\bottomrule
\end{tabular}
\caption{Performance comparison between Qwen3-8B-Reasoning and Qwen3-8B-Reasoning-FTRL-Reinforce++.}
\label{tab:qwen3-8b-reasoning-reinforce++-vs-qwen3-8b-reasoning}
\end{table*}

\begin{table*}[h]
\centering
\begin{tabular}{p{0.45\linewidth}|p{0.45\linewidth}}
\toprule
\rowcolor{gray!10} \multicolumn{1}{c|}{\textbf{Qwen3-8B-Reasoning}} & \multicolumn{1}{c}{\textbf{Qwen3-8B-Non Reasoning}} \\
\midrule
\multicolumn{2}{p{0.9\linewidth}}{\textbf{\textit{Question:}} Where did clashes occur between Palestinians and Israeli police on April 5, 2023?} \\
\multicolumn{2}{p{0.9\linewidth}}{\textbf{\textit{Answer:}} Al-Aqsa Mosque} \\
\midrule
\textbf{\textit{Called Tool:}} event\_locator & \textbf{\textit{Called Tool:}} event\_locator \\
\midrule
\textbf{\textit{Thought:}} ... the event\_locator seems appropriate because it's for locating geopolitical events ... \textcolor{red}{Maybe the tool can handle a general location, like 'Palestinian territories' or `Israel', but I'm not sure.} ... \newline
\textbf{\textit{Filled Parameters:}} \{``event\_type'': ``clash'', ``date'': ``2023-04-05'', ``location'': \textcolor{red}{``Palestinian territories''}, ``verify\_sources'': true, ``source\_credibility\_score'': 0.8, ``event\_severity'': ``high''\} & \textbf{\textit{Filled Parameters:}} \{``keywords'': ``clashes Palestinians Israeli police 2023-04-05'', ``date\_range'': \{``start\_date'': ``2023-04-05'', ``end\_date'': ``2023-04-05''\}, ``topic'': ``conflict'', ``include\_context'': true, ``location'': ``Palestinian territories'', ``significance'': ``national'', ``source\_reliability'': ``high''\} \\
\midrule
\textbf{\textit{Environment Feedback:}} \textcolor{red}{No events found matching the criteria.} & \textbf{\textit{Environment Feedback:}} \textcolor{green!60!black}{Al-Aqsa Mosque. Clashes occurred between Palestinians and Israeli police.} \\
\midrule
\textbf{\textit{Final Output:}} No information was found regarding clashes between Palestinians and Israeli police on April 5, 2023, using the available tools. \textcolor{red}{It is possible that no significant events occurred on this date in the specified context, or the data may not be recorded in the sourced databases.} For the most accurate and up-to-date information, consulting news archives or official reports from the period would be recommended. & \textbf{\textit{Final Output:}} The clashes between Palestinians and Israeli police on April 5, 2023, occurred at the \textcolor{green!60!black}{Al-Aqsa Mosque}. \\
\midrule
\multicolumn{2}{p{0.9\linewidth}}{\textbf{\textit{Analysis:}} In reasoning mode, although Qwen3-8B correctly identified the appropriate tool to use, its reasoning of parameter values lacked accuracy, leading to incorrect entries and failure to retrieve the correct information. This limitation arises because the model's reasoning process is primarily optimized for mathematical tasks, making it poorly suited to the specific demands of tool-use scenarios.}
\\
\midrule
\end{tabular}
\caption{Performance comparison between reasoning and non reasoning modes of Qwen3-8B.}
\label{tab:qwen3-8b-vs-qwen3-8b-reasoning-event-location}
\end{table*}

\clearpage
\onecolumn

\begin{center}
\begin{tcolorbox}[
    colback=white,
    colframe=gray!70!black,
    title=Prompt for Document Generation,
    coltitle=white,
    fonttitle=\bfseries,
    center title,
    rounded corners,
    boxrule=0.6mm,
    width=\linewidth,
    breakable,
    enhanced,
    left=6pt,
    right=6pt,
    top=4pt,
    bottom=4pt
]

\begin{Verbatim}[breaklines=true]
Identify the appropriate tool to solve the given problem and provide an analysis of the tool design. The output should be in JSON format, following the specified structure.

# Steps

1. **Analyze the Problem**: Understand the question and determine the type of information required to answer it.
2. **Tool Design**: Design a tool that can solve the problem, considering the complexity and additional functionalities it might need.
3. **Parameter Specification**: Define the parameters for the tool, ensuring they are comprehensive and flexible for various use cases.
4. **Output Construction**: Format the output in JSON, including both the analysis and the tool schema.

# Notes

- Ensure the tool is versatile enough to handle different but similar queries.
- Consider edge cases.

# Output Format

The output should be a JSON object with the following structure **without any other contents**:
- "analysis": A detailed analysis of the ideas behind the tool design.
- "tool": A JSON schema characterizing the tool, including its name, description, and parameters.

# Example 1

**Question**: What is the walking distance (km) from 8 Oceanside Road to Hope Elementary School?

**Output**:
{{
    "analysis": "The problem involves calculating the walking distance between two locations. This requires a more generalized tool that can handle different types of routes and distances based on the mode of transportation. The tool should integrate a mapping or routing service capable of computing distances for various transport modes, such as pedestrian, cycling, and driving routes. It should also support route preferences like avoiding toll roads or specific types of routes. The design must be flexible enough to allow for diverse transportation methods while providing options to fine-tune distance calculations based on user needs.",
    "tool": {{
        "name": "distance_calculator",
        "description": "A versatile tool to calculate distances between two locations for various modes of transportation (e.g., walking, biking, driving). It provides route-based distances and adjusts for real-world conditions such as road types and traffic.",
        "parameters": {{
            "type": "object",
            "properties": {{
                "origin": {{
                    "type": "string",
                    "description": "Starting point address or coordinates for the distance calculation."
                }},
                "destination": {{
                    "type": "string",
                    "description": "Ending point address or coordinates for the distance calculation."
                }},
                "mode": {{
                    "type": "string",
                    "description": "Mode of transportation to calculate the distance for. Options include 'walking', 'biking', 'driving', etc.",
                    "enum": ["walking", "biking", "driving", "public_transport"]
                }},
                "route_preference": {{
                    "type": "string",
                    "description": "Preferred route type (e.g., shortest, fastest, scenic, etc.). Default is 'shortest'.",
                    "enum": ["shortest", "fastest", "scenic", "avoid_tolls"],
                    "default": "shortest"
                }},
                "unit": {{
                    "type": "string",
                    "description": "Unit of distance to return. Can be 'km', 'miles', or 'meters'. Default is 'km'.",
                    "enum": ["km", "miles", "meters"],
                    "default": "km"
                }},
                "avoid_tolls": {{
                    "type": "boolean",
                    "description": "Indicates whether to avoid toll roads. Default is false.",
                    "default": false
                }},
                "traffic_conditions": {{
                    "type": "string",
                    "description": "Accounts for traffic conditions. Options include 'light', 'moderate', 'heavy'. Default is 'light'.",
                    "enum": ["light", "moderate", "heavy"],
                    "default": "light"
                }},
                "use_pedestrian_routes": {{
                    "type": "boolean",
                    "description": "Whether to prioritize pedestrian paths (e.g., walking or biking routes). Default is false.",
                    "default": false
                }}
            }},
            "required": [
                "origin", "destination", "mode"
            ]
        }}
    }}
}}

# Example 2

**Question**: Which forest is near Mount Everest?

**Output**:
{{
    "analysis": "The problem involves identifying nearby forests to a specific geographical landmark, Mount Everest. This requires a geographical search tool that can pull information based on proximity to a known location. The tool needs to account for various types of forests, their proximity to landmarks, and possibly the specific region around Mount Everest. It should be able to return results from a large database or map of forests and other natural landmarks. The tool should also consider different geographic boundaries, such as countries or regions (e.g., Nepal, Tibet), and support query flexibility for a range of similar queries. Additionally, the tool should handle edge cases such as locations in remote or lesser-documented areas.",
    "tool": {{
        "name": "nearby_forest_locator",
        "description": "A tool designed to identify forests or wooded areas near a specific geographic landmark or location, providing proximity-based results for various regions, including remote or hard-to-reach areas.",
        "parameters": {{
            "type": "object",
            "properties": {{
                "landmark": {{
                    "type": "string",
                    "description": "The name of the landmark or location to search for nearby forests."
                }},
                "radius": {{
                    "type": "number",
                    "description": "The search radius (in kilometers) within which to look for nearby forests. Default is 50 km.",
                    "default": 50
                }},
                "region": {{
                    "type": "string",
                    "description": "The region or country where the search should be focused (e.g., 'Nepal', 'Tibet'). This can help narrow down results."
                }},
                "forest_type": {{
                    "type": "string",
                    "description": "Filter for specific types of forests (e.g., 'tropical', 'boreal', 'temperate'). Optional.",
                    "enum": ["tropical", "boreal", "temperate", "mixed", "dry"]
                }},
                "include_protected": {{
                    "type": "boolean",
                    "description": "Whether to include protected areas or national parks in the search. Default is false.",
                    "default": false
                }}
            }},
            "required": ["landmark"]
        }}
    }}
}}

**Question**: {question}

**Output**:
\end{Verbatim}

\end{tcolorbox}
\end{center}
\captionof{figure}{The prompt used for document generation.}
\label{fig:prompt-dg}

\begin{center}
\begin{tcolorbox}[
    colback=white,
    colframe=gray!70!black,
    title=Prompt for Function Integration,
    coltitle=white,
    fonttitle=\bfseries,
    center title,
    rounded corners,
    boxrule=0.6mm,
    width=\linewidth,
    breakable,
    enhanced,
    left=6pt,
    right=6pt,
    top=4pt,
    bottom=4pt
]

\begin{Verbatim}[breaklines=true]
Your task is to analyze a set of tool documents in JSON schema format to identify any tools with the same functionality and merge them if needed. Retain the functionality from each separate document, ensure the final merged document remains fully compatible, and then output your reasoning and the merged documents strictly in the specified JSON structure.

# Steps

1. **Parse and Understand**: Begin by parsing each tool document's JSON schema to understand its functionality, inputs, and outputs. Identify key features that define its purpose and operations.

2. **Compare Documents**: Systematically compare each document to identify tools with identical or overlapping functionalities. Look for description of each tool to determine similarities.

3. **Merge Tools**: For each group of functionally identical tools, merge them into a single new schema. Ensure the merged schema accommodates all functionalities from the original tools without loss of essential detail or compatibility.

4. **Compose Analysis**: Draft your reasoning process, describing how the schemas were compared, how conclusions on identical functionalities were reached, and details of how they were merged.

# Output Format

Your output must be valid JSON according to the following structure:

- `"analysis"`: A string detailing your reasoning, including how you compared schemas, identified identical functionalities, and performed merges.
  
- `"merged"`: An array of objects, each containing:
  - `"id"`: A list of tool numbers (in ascending order) that you decided to merge.
  - `"document"`: The merged document if applicable.
  - Tools that do not need to be merged must not be displayed here.
  - If no tools need merging, use `"merged": null`.

# Notes

- Ensure that the final merged document is fully compatible and retains all original functionalities.
- Use clear reasoning and specify the parameters or features that led to the decision to merge the tools.
- Maintain the integrity and accuracy of the original tool documents in the merged results.

**Tool Documents**:
{documents}

**Output**:
\end{Verbatim}

\end{tcolorbox}
\end{center}
\captionof{figure}{The prompt used for function integration.}
\label{fig:prompt-fi}

\clearpage

\begin{center}
\begin{tcolorbox}[
    colback=white,
    colframe=gray!70!black,
    title=Prompt for Complexity Scaling,
    coltitle=white,
    fonttitle=\bfseries,
    center title,
    rounded corners,
    boxrule=0.6mm,
    width=\linewidth,
    breakable,
    enhanced,
    left=6pt,
    right=6pt,
    top=4pt,
    bottom=4pt
]

\begin{Verbatim}[breaklines=true]
Refine the design of a tool by enhancing its description and increasing the complexity of parameters, while maintaining compatibility with the original functionality.

# Steps

1. **Analyze the Current Tool**: Examine the existing tool's description and parameters to understand its functionality and limitations.
2. **Identify Areas for Refinement**: Determine which aspects of the tool can be improved or expanded to better meet real-world requirements.
3. **Refine the Description**: Refine existing parameters so that each parameter value is an objective entity. Introduce new parameters to increase complexity and utility, but ensure full compatibility with legacy functionality.
4. **Ensure Compatibility**: Verify that the refined version remains compatible with the original tool's purpose and structure.

# Output Format

The output should be in JSON format with the following structure **without any other contents**:
- "analysis": Analysis of ideas about refining the tool.
- "refined_version": The version after refinement, should be follow JSON SCHEMA format as the original tool.

# Notes

- Ensure that any new parameters added are relevant and enhance the tool's functionality.
- Maintain backward compatibility with the original tool's design and purpose.

**Tool**:
{tool}

**Output**:
\end{Verbatim}

\end{tcolorbox}
\end{center}
\captionof{figure}{The prompt used for complexity scaling.}
\label{fig:prompt-cs}

\clearpage

\begin{center}
\begin{tcolorbox}[
    colback=white,
    colframe=gray!70!black,
    title=Prompt for Localized Deployment,
    coltitle=white,
    fonttitle=\bfseries,
    center title,
    rounded corners,
    boxrule=0.6mm,
    width=\linewidth,
    breakable,
    enhanced,
    left=6pt,
    right=6pt,
    top=4pt,
    bottom=4pt
]

\begin{Verbatim}[breaklines=true]
Create a function implementation based on a provided tool document and question-answer pairs. The implementation should strictly adhere to the tool's specifications and include robust error handling.

# Steps

1. **Understand the Tool Document**: Carefully review the tool document to identify the function name, parameter names, and types. Ensure that these details are used as-is in the function implementation.
2. **Analyze Question-Answer Pairs**: Examine these pairs to understand how questions map to function inputs and how answers should be derived from function outputs.
3. **Implement the Function**:
   - Use the tool-specified function name.
   - Define parameters exactly as specified in the tool document.
   - Implement logic to correctly derive answers for questions based on the input parameters.
   -  When parameters are assigned default values, Make sure that the function return value contains the complete given answer, i.e., the answer is a substring of the return value.
   - Ensure the function is capable of returning various responses, simulating additional return values if necessary.
4. **Error Handling**: Develop a comprehensive mechanism to return error messages for incorrect inputs or other issues, ensuring the function operates reliably in all scenarios.

# Output Format

The result should be output in JSON format, adhering to the following structure **without anything else**: 
- "analysis": A detailed explanation of the function design, including reasoning for parameter choices and exception handling logic.
- "function": The specific implementation of the function, with code and comments that describe each component.

# Notes

- Match parameter names and types exactly with those specified in the tool document.
- Simulate additional return values as needed, based on the tool's documentation.
- Rely only on Python3's built-in libraries for all function implementations.
- Ensure comprehensive error handling to anticipate and manage potential issues effectively.

**Tool Document**:
{document}

**Question-Answer Pairs**: 
{pairs}

**Output**:
\end{Verbatim}

\end{tcolorbox}
\end{center}
\captionof{figure}{The prompt used for localized deployment.}
\label{fig:prompt-ld}

\begin{center}
\begin{tcolorbox}[
    colback=white,
    colframe=gray!70!black,
    title=Prompt for Code Verify,
    coltitle=white,
    fonttitle=\bfseries,
    center title,
    rounded corners,
    boxrule=0.6mm,
    width=\linewidth,
    breakable,
    enhanced,
    left=6pt,
    right=6pt,
    top=4pt,
    bottom=4pt
]

\begin{Verbatim}[breaklines=true]
Analyze the relationship between a given piece of function code and a problem to provide the appropriate function call to solve the problem, ensuring that all parameter values are derived from the problem.

# Steps

1. **Understand the Problem**: Read and comprehend the details of the given problem.
2. **Analyze the Code**: Examine the provided function code to ascertain how it addresses the problem.
3. **Confirm Code-to-Problem Suitability**: Determine if the function correctly solves the problem as described.
4. **Derive Function Call**: Craft a function call using the problem's specific details for parameter values.

# Output Format

Output the result in the following JSON format without any additional text:

- "analysis": A description analyzing how the given code relates to and addresses the problem.
- "call": The function call formatted as func(param="value"), using the problem's details for the parameter values.
  - If the fucntion can not used to solve the problem, return "call": null.

# Examples

**Example 1:**

- Input Problem: "Calculate the area of a rectangle with width=5 and height=10."
- Input Code: `def calculate_area(width, height): return width * height`

- JSON Output:
  - {{
    "analysis": "The function calculate_area computes the area of a rectangle by multiplying width and height. Given the problem, it appropriately solves it.",
    "call": "calculate_area(width=5, height=10)"
  }}

**Example 2:**

- Input Problem: "Find the square of the number 7."
- Input Code: `def square(num): return num * num`

- JSON Output:
  - {{
    "analysis": "The function square returns the square of the input number. It is suitable for finding the square of 7 as described in the problem.",
    "call": "square(num=7)"
  }}

# Notes

- Ensure the function call uses the exact parameter names as defined in the function code.
- The analysis should concisely justify the relationship between the function and the problem solution.

- Input Problem: {question}
- Input Code: {code}

- JSON Output:
\end{Verbatim}
\end{tcolorbox}
\end{center}
\captionof{figure}{The prompt used for code verify.}
\label{fig:prompt-cv}

\end{document}